\documentclass[final,12pt]{colt2023} 

\title[Non-stationary Projection-free Online Learning]{Non-stationary Projection-free Online Learning \\ with Dynamic and Adaptive Regret Guarantees}
\usepackage{times}

\newcommand{\RNum}[1]{\uppercase\expandafter{\romannumeral #1\relax}}

\usepackage{makecell}
\usepackage{multirow}

\usepackage{mathtools}
\usepackage{amssymb}
\usepackage{amsmath}
\usepackage{soul}
\usepackage{xcolor}
\usepackage{algorithm}
\usepackage{algorithmic}
\usepackage{booktabs}

\usepackage{hyperref}

\usepackage{graphicx}
\usepackage{appendix}
\usepackage{paralist}

\newtheorem{assumption} {Assumption}

\def \bx        {\mathbf{x}}
\def \by        {\mathbf{y}}
\def \bz        {\mathbf{z}}
\def \bv        {\mathbf{v}}

\def \bu        {\mathbf{u}}
\def \be        {\mathbf{e}}

\def \Tilby     {\Tilde{\by}}
\def \Tileta    {\Tilde{\eta}}

\def \nabft     {\nabla f_t}
\def \nabf      {\nabla f}

\def \argmin    {\mathop{\arg\min}}
\def \argmax    {\mathop{\arg\max}}

\def \A         {\mathcal{A}}
\def \B         {\mathcal{B}}
\def \C         {\mathcal{C}}

\def \H         {\mathcal{H}}
\def \I         {\mathcal{I}}
\def \O         {\mathcal{O}}
\def \K         {\mathcal{K}}

\coltauthor{%
 \Name{Yibo Wang} \Email{wangyb@lamda.nju.edu.cn}\\
 \Name{Wenhao Yang} \Email{yangwh@lamda.nju.edu.cn}\\
 \Name{Wei Jiang} \Email{jiangw@lamda.nju.edu.cn}\\
 \addr National Key Laboratory for Novel Software Technology, Nanjing University, Nanjing 210023, China 
 \AND
 \Name{Shiyin Lu} \Email{running.lsy@alibaba-inc.com}\\
 \Name{Bing Wang} \Email{lingfeng.wb@alibaba-inc.com}\\
 \Name{Haihong Tang} \Email{piaoxue@taobao.com}\\
 \addr Alibaba Group, Hangzhou 310052, China 
 \AND
 \Name{Yuanyu Wan} \Email{wanyy@zju.edu.cn}\\
 \addr School of Software Technology, Zhejiang University, Ningbo 315048, China
 \AND
 \Name{Lijun Zhang} \Email{zhanglj@lamda.nju.edu.cn}\\
 \addr National Key Laboratory for Novel Software Technology, Nanjing University, Nanjing 210023, China 
}

\begin{document}

\maketitle

\begin{abstract} 
Projection-free online learning has drawn increasing interest due to its efficiency in solving high-dimensional problems with complicated constraints. However, most existing projection-free online methods focus on minimizing the static regret, which unfortunately fails to capture the challenge of changing environments. In this paper, we investigate non-stationary projection-free online learning, and choose  \emph{dynamic regret} and \emph{adaptive regret} to measure the performance. Specifically, we first provide a novel dynamic regret analysis for an existing projection-free method named $\text{BOGD}_\text{IP}$, and establish an $\mathcal{O}(T^{3/4}(1+P_T))$ dynamic regret bound, where $P_T$ denotes the path-length of the comparator sequence. Then, we improve the upper bound to  $\mathcal{O}(T^{3/4}(1+P_T)^{1/4})$ by running multiple  $\text{BOGD}_\text{IP}$ algorithms with different step sizes in parallel, and tracking the best one on the fly.  Our results are the first general-case dynamic regret bounds for projection-free online learning, and can recover the existing $\mathcal{O}(T^{3/4})$ static regret by setting $P_T = 0$. Furthermore, we propose a projection-free method to attain an $\tilde{\mathcal{O}}(\tau^{3/4})$ adaptive regret bound for any interval with length $\tau$, which nearly matches the static regret over that interval. The essential idea is to maintain a set of $\text{BOGD}_\text{IP}$ algorithms dynamically, and combine them by a  meta algorithm. Moreover, we demonstrate that it is also equipped with an $\mathcal{O}(T^{3/4}(1+P_T)^{1/4})$ dynamic regret bound. Finally, empirical studies verify our theoretical findings.
\end{abstract}

\section{Introduction} \label{sec:intro}

In many online learning problems, the decision constraint sets are often high-dimensional and complicated, rendering  optimization over such sets challenging. In these cases, traditional projection-based methods, such as Online Gradient Descent (OGD) \citep{ICML:2003:Zinkevich}, often suffer heavy computational costs due to the time-consuming or even intractable projection operations. To address this limitation, projection-free online methods, which replace projections with less expensive computations (e.g., linear optimizations) and thus can be  implemented efficiently in many cases of interest, have drawn considerable attention in the online learning community \citep{ICML:2012:Hazan,Others:2016:Garber,NeurIPS:2016:Huang,AISTATS:2019:Levy,COLT:2020:Hazan,ArXiv:2020:Molinaro,Others:2021:Kalhan,AAAI:2021:Wan,AAAI:2021:Wan:B,AISTATS:2021:Garber,COLT:2022:Garber,COLT:2022:Mhammedi,ALT:2023:Lu,ArXiv:2023:Wan,ArXiv:2023:Garber}. 

The studies of projection-free online methods follow the framework of Online Convex Optimization (OCO), which can be regarded as a repeated game between a learner against an adversary \citep{Others:2012:Shalev-Shwartz}. At round $t$, the learner chooses an action $\bx_t$ from a convex domain set $\K$, and then suffers an instantaneous loss $f_t(\bx_t)$, where the convex loss function  $f_t(\cdot):\K \rightarrow \mathbb{R}$ is chosen by the adversary. 
The majority of existing projection-free methods, e.g.,~Online Frank-Wolfe (OFW) \citep{ICML:2012:Hazan}, minimize the static regret:
\begin{equation}
    \label{eq:static_regret}
     \text{{\rm Regret}}_T  = \sum_{t=1}^T f_t(\bx_t) -  \min_{\bx \in \K} \sum_{t=1}^T f_t(\bx), 
\end{equation}
which benchmarks the cumulative loss of the online method against that of the best fixed action in hindsight.
However, in real-world scenarios such as online recommendation and online traffic scheduling \citep{Others:2016:Hazan}, this static metric is unsuitable as the environments are non-stationary and the best action is drifting over time. To tackle this issue, two novel performance metrics: dynamic regret and adaptive regret, are proposed independently \citep{ICML:2003:Zinkevich,ECCC:2007:Hazan,ICML:2015:Daniely}.

The dynamic regret stems from \citet{ICML:2003:Zinkevich}, who defines  
\begin{equation}
    \label{eq:general_dynamic_regret}
    \text{{\rm D-Regret}}_T(\bu_1, \cdots, \bu_T)  = \sum_{t=1}^T f_t(\bx_t) - \sum_{t=1}^T f_t(\bu_t),
\end{equation}
where $\bu_1, \cdots, \bu_T \in \K$ are any possible comparators.
Unfortunately, obtaining a sublinear dynamic regret with arbitrarily varying sequences is impossible. As a result, to establish a meaningful bound, it is common to introduce some regularities of the comparator sequence, such as the path-length
\begin{equation*}
    P_T = \sum_{t=2}^T \|\bu_{t-1} - \bu_{t}\|_2.
\end{equation*}

The adaptive regret is originally introduced by \citet{ECCC:2007:Hazan}, and further strengthened by \citet{ICML:2015:Daniely}. Formally, it is defined as
\begin{equation}
    \label{eq:strong_adaptive_regret}
    \begin{split}
        &\text{{\rm SA-Regret}}_T(\tau) =\max_{[s, s + \tau - 1] \subseteq [T]} \left\{  \sum_{t=s}^{s + \tau - 1} f_t(\bx_t) - \min_{\bx \in \K}\sum_{t=s}^{s + \tau - 1} f_t(\bx)  \right\}, 
    \end{split}
\end{equation}
which is  the maximum  static regret over any interval with the length $\tau$. Since in different intervals the best actions can be different, \eqref{eq:strong_adaptive_regret}  essentially measures the performance  of the online method against changing comparators.

In the literature, only a few projection-free online methods  \citep{Others:2021:Kalhan,AAAI:2021:Wan:B,ArXiv:2023:Wan} have investigated   dynamic regret minimization, but all of them focus on the worst case of \eqref{eq:general_dynamic_regret}, where $\bu_t \in \argmin_{\bu \in \K} f_t(\bu)$ is a minimizer of $f_t(\cdot)$. However, the worst-case dynamic regret is too pessimistic, and  cannot recover the static regret bound of   previous methods \citep{ICML:2012:Hazan,COLT:2020:Hazan}. Besides, there exist two studies \citep{COLT:2022:Garber,ALT:2023:Lu} that propose projection-free methods for adaptive regret minimization. However, \citet{COLT:2022:Garber} only consider a weak form of \eqref{eq:strong_adaptive_regret} which  does not respect short intervals well, and the method of \citet{ALT:2023:Lu} could be time-consuming in many popular domains, e.g.,~bounded trace norm matrices and matroid polytopes \citep{COLT:2022:Mhammedi}.

In this paper, we choose  \eqref{eq:general_dynamic_regret} and  \eqref{eq:strong_adaptive_regret} as the performance metrics, and propose two novel methods for non-stationary projection-free online learning. Specifically, in the dynamic regret minimization,  we first establish a novel dynamic regret bound of $\O(T^{3/4}(1+P_T))$ for an existing projection-free variant of Online Gradient Descent, termed as $\text{BOGD}_\text{IP}$ \citep{COLT:2022:Garber}.\footnote{In \citet{COLT:2022:Garber},  $\text{BOGD}_\text{IP}$ is referred to as Blocked Online Gradient Descent with Linear Optimization Oracle (LOO-BOGD).} Then, we improve the upper bound to $\O(T^{3/4}(1+P_T)^{1/4})$ by proposing a two-layer method named POLD, which  maintains multiple  $\text{BOGD}_\text{IP}$ algorithms with different step sizes, and tracks the best one on the fly by a meta algorithm. 
In the adaptive regret minimization,  we propose a novel projection-free method named POLA, which attains an $\Tilde{\O}(\tau^{3/4})$ adaptive regret bound for any interval with the length $\tau$. 
The key idea is to construct a set of intervals dynamically, run a  $\text{BOGD}_\text{IP}$ algorithm that aims to minimize the static regret for each interval, and combine them by a meta algorithm.
Moreover, we show that our POLA can also minimize the dynamic regret, and ensures an $\O(T^{3/4}(1+P_T)^{1/4})$ bound. Notably, although POLA  can achieve the same dynamic regret bound as POLD, the latter one is still valuable in the sense that it employs a clearer structure and a simpler meta algorithm, rendering it much easier to comprehend and implement.

\paragraph{Contributions.} 
We summarize the contributions of this work below.

\begin{compactitem} 
    \item  For dynamic regret, we  first provide a novel analysis for $\text{BOGD}_\text{IP}$ \citep{COLT:2022:Garber}, and establish an $\O(T^{3/4}(1+P_T))$ dynamic regret. Then, we improve this bound to $\O(T^{3/4}(1+P_T)^{1/4})$ by proposing a two-layer  method  named POLD. Note that the obtained bounds can recover the previous $\O(T^{3/4})$  static regret  \citep{ICML:2012:Hazan} by setting $P_T=0$. To the best of our knowledge, these are the \textit{first} general-case dynamic regret bounds in projection-free online learning.
    \item For adaptive regret, based on  $\text{BOGD}_\text{IP}$,  we propose a novel projection-free method named POLA  and obtain an $\Tilde{\O}(\tau^{3/4})$ adaptive regret which nearly matches previous static results.  Moreover, we show that POLA can also ensure an $\O(T^{3/4}(1+P_T)^{1/4})$ dynamic regret bound. In other words, it can minimize  dynamic regret and adaptive regret simultaneously.
    \item We conduct experiments  on practical problems to verify our theoretical findings in dynamic regret and adaptive regret minimization. Empirical results  demonstrate the advantage of proposed methods.
\end{compactitem}

\section{Related work}

In this section, we briefly review related work in dynamic regret and adaptive regret.

\subsection{Dynamic regret}
In the literature, dynamic regret has two different forms.
One is the general case  \eqref{eq:general_dynamic_regret} introduced by \citet{ICML:2003:Zinkevich}, who defines it as the difference between the cumulative loss of the online method and that of \emph{any} possible comparator sequence. In this seminal work, \citet{ICML:2003:Zinkevich} establishes the first general-case bound of $\O(\sqrt{T}(1+P_T))$ for OGD. 
Later,  \citet{NeurIPS:2018:Zhang} improve the upper bound to $\O(\sqrt{T(1+P_T)})$, motivated by the strategy of maintaining multiple step sizes in MetaGrad \citep{NeurIPS:2016:Erven,ICML:2019:Mhammedi,JMLR:2021:Erven}.
In recent years,  several studies have further investigated the general-case dynamic regret by leveraging the curvature of loss functions, such as exponential concavity \citep{COLT:2021:Baby}  and strong convexity \citep{AISTATS:2022:Baby}.

The other is the worst case of \eqref{eq:general_dynamic_regret}, which specializes the comparators as the  minimizers of loss functions \citep{Others:2015:Besbes,AISTATS:2015:Jadbabaie,CDC:2016:Mokhtari,ICML:2016:Yang,NeurIPS:2019:Baby}:
\begin{equation}
    \label{eq:worst_dynamic_regret}
    \text{{\rm D-Regret}}_T(\bu^*_1, \cdots, \bu^*_T)  = \sum_{t=1}^T f_t(\bx_t) - \sum_{t=1}^T  f_t(\bu^*_t)
\end{equation}
where $\bu_t^* \in \argmin_{\bu \in \K} f_t(\bu)$ is a minimizer of $f_t(\cdot)$.
However, as pointed out by \citet{NeurIPS:2018:Zhang}, the worst-case dynamic regret \eqref{eq:worst_dynamic_regret} is too pessimistic  and could lead to overfitting in the stationary problems.

In projection-free online learning, \citet{Others:2021:Kalhan} and \citet{AAAI:2021:Wan:B,ArXiv:2023:Wan} have investigated the dynamic regret recently, but they only consider the worst-case formulation \eqref{eq:worst_dynamic_regret}. Specifically,
for  smooth and convex losses, \citet{Others:2021:Kalhan}  establish an $\O(\sqrt{T} (1 + F_T + \sqrt{D_T}))$ worst-case bound, where $F_T = \sum_{t=2}^T \sup_{\bx \in \K} |f_{t}(\bx) - f_{t-1}(\bx)|$  and $D_T = \sum_{t=2}^T \|\nabf_t (\bx_{t}) - \nabf_{t-1}(\bx_{t-1})\|_2^2$.
For convex losses and strongly convex losses, \citet{AAAI:2021:Wan:B} develop  the
$\O(\max\{T^{2/3}F_T^{1/3}, \sqrt{T}\})$ and $\O(\max\{\sqrt{T F_T\log T}, \log T\})$   worst-case  bounds, respectively. Very recently, \citet{ArXiv:2023:Wan} refine the analysis of \citet{Others:2021:Kalhan}, achieving an improved $\O(\sqrt{T(1+F_T)})$ bound.
However, due to the weakness of \eqref{eq:worst_dynamic_regret}, their bounds can be very loose for any other comparators, and  cannot recover the static regret of existing methods, e.g.,~$\O(T^{3/4})$ for convex losses \citep{ICML:2012:Hazan}.

\begin{table}[t]
    \centering
    
    \caption{ 
        Summary of  existing methods in non-stationary projection-free online learning. Abbreviations: linear optimization $\rightarrow$ LO, membership operation $\rightarrow$ MO,  worst-case dynamic regret \eqref{eq:worst_dynamic_regret} $\rightarrow$ WD-R,  general-case dynamic regret \eqref{eq:general_dynamic_regret} $\rightarrow$ D-R,  weak adaptive regret \eqref{eq:weak_adaptive_regret} $\rightarrow$ A-R,    strongly adaptive regret \eqref{eq:strong_adaptive_regret} $\rightarrow$ SA-R.
    }
    \setlength\tabcolsep{1.5pt}
    \begin{tabular}{ccccc}
           \toprule
           Method &  \makecell[c]{Loss } & Operation &    Metric &  Bound\\
           \midrule
           \citet{Others:2021:Kalhan} & smooth \& convex & LO &   WD-R  & $\O(\sqrt{T} (1 + F_T + \sqrt{D_T}))$  \\
           \citet{ArXiv:2023:Wan}  & smooth \& convex & LO &  WD-R  & $\O(\sqrt{T(1+F_T)})$ \\
           \multirow{2}{*}{\citet{AAAI:2021:Wan:B}}  
           & convex & LO &   WD-R  &  $\O(\max\{T^{2/3}F_T^{1/3}, \sqrt{T}\})$  \\
           & strongly convex & LO &   WD-R & $\O(\max\{\sqrt{T F_T\log T}, \log T\})$  \\
           \textbf{$\text{BOGD}_\text{IP}$}  (this work) & convex & LO & D-R   &  $\O(T^{3/4}(1+P_T))$\\
           \textbf{POLD} (this work) & convex & LO &  D-R  &  $\O(T^{3/4}(1+P_T)^{1/4})$\\
           \textbf{POLA} (this work) & convex & LO &  D-R  &  $\O(T^{3/4}(1+P_T)^{1/4})$\\
           \hline
           \citet{COLT:2022:Garber} & convex & LO & A-R &  $\mathcal{O}(T^{3/4})$ \\
           \citet{ALT:2023:Lu} & convex &  MO & SA-R & $\Tilde{\O} (\sqrt{\tau})$ \\
           \textbf{POLA} (this work) & convex & LO &  SA-R   &  $\Tilde{\O}(\tau^{3/4})$\\
           \bottomrule
    \end{tabular}
    \label{tab:related_work}
\end{table}

\subsection{Adaptive regret}
Prior work in adaptive regret minimization mainly focus on the setting of Prediction with Expert Advice (PEA) \citep{Others:1994:Littlestone,STOC:1997:Freund,Others:2012:Gyorgy,COLT:2015:Luo,JMLR:2016:Adamskiy}, and  OCO \citep{ECCC:2007:Hazan,ICML:2015:Daniely,AISTATS:2017:Jun,Others:2017:Jun,ICML:2019:Zhang}. In this section, we specifically introduce the related work of the latter one.

\citet{ECCC:2007:Hazan} first introduce the notion of adaptive regret, but in a weak form:
\begin{equation}
    \label{eq:weak_adaptive_regret}
    \text{{\rm A-Regret}}_T = \max_{[s, e] \subseteq [T]} \left\{  \sum_{t=s}^e f_t(\bx_t) - \min_{\bx \in \K} \sum_{t=s}^e f_t(\bx)  \right\}, 
\end{equation}
which is the maximum static regret over any contiguous interval. To minimize \eqref{eq:weak_adaptive_regret}, they propose Follow the Leading History (FLH) with an $\O(d \log^2 T)$ weak adaptive regret bound for exponentially concave losses where $d$ denotes the dimensionality. 
However, \eqref{eq:weak_adaptive_regret} could be dominated by long intervals and hence, cannot respect short intervals well. For example, one may obtain an $\O(\sqrt{T})$ weak adaptive regret for OGD, but this is vacuous for the intervals with length $o(\sqrt{T})$ \citep{Others:2016:Hazan}. 
For this reason, \citet{ICML:2015:Daniely} put forth  the (strongly) adaptive regret \eqref{eq:strong_adaptive_regret}, 
and design a two-layer algorithm named Strongly Adaptive Online Learner (SAOL). 
The basic idea is first to construct  a set of Geometric Covering (GC) intervals and  for each interval, run an OGD algorithm that can obtain the optimal static regret. Then, SAOL  combines the actions of  these OGD algorithms by a meta algorithm. We observe that the technique of constructing GC intervals can  be traced back to the prior studies \citep{Others:1997:Willems,Others:2012:Gyorgy}.

In projection-free online learning,  \citet{COLT:2022:Garber} study the weak version of adaptive regret \eqref{eq:weak_adaptive_regret},  and propose a projection-free extension of OGD named $\text{BOGD}_\text{IP}$ with an $\O(T^{3/4})$  bound. Unfortunately, due to the limitation of  \eqref{eq:weak_adaptive_regret}, their bound does not respect short intervals well. Very recently, following the framework of  SAOL, \citet{ALT:2023:Lu} propose a novel two-layer method to minimize \eqref{eq:strong_adaptive_regret}. Different from previous projection-free algorithms, e.g.,~OFW \citep{ICML:2012:Hazan}, their method circumvents the projections with  membership operations \citep{COLT:2022:Mhammedi}. However, such operations could be inefficient in many practical scenarios, e.g.,~bounded trace norm matrices and matroid polytopes \citep{COLT:2022:Mhammedi}. Besides, in each round, their method need to perform $\O(\log T)$ membership operations for each expert algorithm, which brings heavy computational costs when $T$ is large.

\paragraph{Summary.}
While a few work have investigated  non-stationary projection-free online learning (see Table \ref{tab:related_work} for details),  they are still unsatisfactory in the following aspects:

\begin{compactitem}
     \item In the dynamic regret minimization, there is no study optimizing the general-case form \eqref{eq:general_dynamic_regret}, which is more challenging since it needs to build a universal guarantee over any comparator sequences.

    \item In the adaptive regret minimization, although \citet{ALT:2023:Lu} have established bounds for $\eqref{eq:strong_adaptive_regret}$, their method is based on the membership operations, instead of the more popular linear optimizations.
\end{compactitem}

\section{Main results}
In this section, we first introduce the basic assumptions. Then, we present our proposed methods as well as their theoretical guarantees in dynamic regret and adaptive regret minimization.

\subsection{Assumptions}
Similar to previous studies on OCO, we adopt the following standard assumptions  \citep{Others:2012:Shalev-Shwartz,Others:2016:Hazan}.

\begin{assumption}
    \label{assump:K-bound}
    The convex decision set $\mathcal{K}$ contains the origin $\mathbf{0}$, and  belongs to an Euclidean ball $R\B$ with the diameter $D = 2R$, i.e.,
    \begin{equation}
        \label{eq:K-bound}
        \forall \bx, \bx' \in \mathcal{K}, \|\bx - \bx'\|_2 \leq D.
    \end{equation}

\end{assumption}
\begin{assumption}
    \label{assump:Lipschitz}
    At each round $t$, the loss function $f_{t}(\cdot)$ is $G$-Lipschitz over $\mathcal{K}$, i.e.,
    \begin{equation}
        \label{eq:Lipschitz}
        \forall \bx, \by \in \mathcal{K}, |f_{t}(\bx) - f_{t}(\by)| \leq G \| \bx - \by\|_2.
    \end{equation}
\end{assumption}

\begin{assumption}
    \label{assump:convex}
    At each round $t$, the loss function $f_{t}(\cdot)$ is convex over $\mathcal{K}$, i.e.,
    \begin{equation}
        \label{eq:convex_function}
        \forall \bx, \by \in \mathcal{K},  f_t(\by) \geq f_t(\bx) + \nabla f_t(\bx)^\top (\by - \bx).
    \end{equation}
\end{assumption}

\begin{assumption}
    \label{assump:f-bound}
    At each round $t$, the loss function value $f_{t}(\bx)$ belongs to $[0, 1]$  for any $\bx \in \mathcal{K}$, i.e.,
    \begin{equation}
        \forall \bx \in \mathcal{K}, ~ 0 \leq f_t(\bx) \leq 1.
    \end{equation}
\end{assumption}

\subsection{Projection-free dynamic regret}

We first revisit $\text{BOGD}_{\text{IP}}$ \citep{COLT:2022:Garber}, of which  the key idea is to replace the projection operation  with an infeasible projection oracle $\O_{\text{IP}}$,  defined as following.
\begin{definition}
       Let $\O_{\text{IP}}$ be an infeasible projection oracle over $\K \subseteq R \B$, and $\epsilon$ be the error tolerance. Then, for any input points $(\bx_0, \by_0) \in \K \times \mathbb{R}^d$, the infeasible projection oracle returns
       \begin{equation*}
              \bx, \Tilby  =  \mathcal{O}_{\text{IP}}(\mathcal{K}, \epsilon, \bx_0, \by_0),
       \end{equation*}
       where $(\bx, \Tilby) \in \K \times R \B$, and $\|\bx - \Tilby\|_2 \leq \sqrt{3 \epsilon}$ and  $\forall \bz \in \K, \|\Tilby  - \bz\|_2 \leq \|\by_0  - \bz\|_2$.
\end{definition}
\textbf{Remark:} $\O_{\text{IP}}$ can be implemented efficiently by solving linear optimizations. We briefly introduce this implementation in Appendix \ref{sec:IP}, and refer  interested readers to  \citet{COLT:2022:Garber} for a deeper comprehension.

Besides, $\text{BOGD}_{\text{IP}}$ utilizes  the blocking technique \citep{AISTATS:2020:Garber,COLT:2020:Hazan}, which divides the time horizon $T$ into equally-sized blocks and only conducts updating at the end of each block. In other words, for each block $m$, $\text{BOGD}_{\text{IP}}$ maintains  $(\bx_m, \Tilby_m)  \in \K \times R\B$, and updates them at the last round of block $m$. 
To be precise,  $\text{BOGD}_{\text{IP}}$ first performs gradient descent on $\Tilby_m$ with the step size $\eta$:
\begin{equation}
    \label{eq:LOO-BOGD_1}
    \by_{m+1} = \Tilby_{m} - \eta \sum_{r=(m-1)K+1}^{mK} \nabla  f_r(\bx_{m}),
\end{equation}
where $K$ is the block size and $\sum_{r=(m-1)K+1}^{mK} \nabla  f_r(\bx_{m})$ is the sum of all gradients during the block $m$. 
Then, $\text{BOGD}_{\text{IP}}$ invokes $\O_{\text{IP}}$ to obtain $\bx_{m+1}$ and $\Tilby_{m+1}$ for the next block:
\begin{equation}
    \label{eq:LOO-BOGD_2}
    \bx_{m+1}, \Tilby_{m+1} =  \mathcal{O}_{\text{IP}}(\mathcal{K}, \epsilon, \bx_{m}, \by_{m+1}).
\end{equation}
With  appropriate parameters, we can prove that $\text{BOGD}_{\text{IP}}$ requires $\O(T^{1/2})$ invocations of $\O_{\text{IP}}$, and each invocation solves $\O(T^{1/2})$ linear optimizations. As a result, there are at most $\O(T)$ linear optimizations for the time horizon $T$. 
We summarize the detailed procedure in Algorithm \ref{alg2}.

In the prior study, \citet{COLT:2022:Garber} have investigated  the weak adaptive regret \eqref{eq:weak_adaptive_regret}.  Different from them, we focus on minimizing the general-case dynamic regret \eqref{eq:general_dynamic_regret} and establish an $\O(T^{3/4}  ( 1 + P_T ))$ bound for $\text{BOGD}_\text{IP}$ as shown in Theorem \ref{theorem:dynamic_BOGD}. The intuition lies in that  $\text{BOGD}_\text{IP}$ is a projection-free variant of OGD, which is very suitable for dynamic regret minimization \citep{ICML:2003:Zinkevich}.

\begin{algorithm}[tb]
    \caption{\underline{B}locked \underline{O}nline \underline{G}radient \underline{D}escent with \underline{I}nfeasible \underline{P}rojections ($\text{BOGD}_{\text{IP}}$)}
    \label{alg2}
    \textbf{Input}: Number of rounds $T$, domain set $\mathcal{K}$, step size $\eta$, infeasible projection oracle $\mathcal{O}_{\text{IP}}$
		
    \textbf{Initialization}: Choose arbitrary point $\bx_1 \in \mathcal{K}$ and set $\Tilby_1 = \bx_1, m = 1$, block size $K = \eta^{-2/3}$ and error tolerance $\epsilon = \eta^{2/3}$.
	
    \begin{algorithmic}[1]
		
        \FOR{$t=1$ to $T$}

        \STATE  Submit $\bx_t = \bx_m$, observe $f_t(\bx_{t})$ and obtain $\nabla f_t(\bx_{t})$

        \IF {$t \mod K = 0$ }

        \STATE  Update $\by_{m+1}$ according to \eqref{eq:LOO-BOGD_1}

        \STATE  Set $\bx_{m+1}, \Tilby_{m+1}$ according to \eqref{eq:LOO-BOGD_2}, and  $m = \lfloor t / K \rfloor + 1$

        \ENDIF

        \ENDFOR

    \end{algorithmic}
\end{algorithm}

\begin{theorem}
    \label{theorem:dynamic_BOGD}
    Let $\eta = T^{-3/4}$, $K = \eta^{-2/3} = T^{1/2}$ and $\epsilon = \eta^{2/3} = T^{-1/2}$. Under Assumptions \ref{assump:K-bound}, \ref{assump:Lipschitz} and \ref{assump:convex}, Algorithm \ref{alg2} guarantees
    \begin{equation*}
        \begin{split}
            \text{{\rm D-Regret}}_T(\bu_1, \cdots, \bu_T) &\leq \O\left( \eta^{1/3}T + \eta^{-1} \left( 1 + P_T\right) \right) 
            = \mathcal{O}\left( T^{3/4} \left( 1 + P_T\right)\right).
        \end{split}
    \end{equation*}
    Moreover, the overall number of solving linear optimizations is $ \mathcal{O}(T)$.
\end{theorem}
\textbf{Remark:} Our result is the first general-case dynamic regret bound in projection-free online learning, and can automatically adapt to the nature of environments. For example, when the comparators are fixed (i.e.,~$P_T = 0$), our dynamic regret degenerates to $\O(T^{3/4})$, which matches the static regret bound of \citet{ICML:2012:Hazan}.
To be specific, we have the following corollary, which can also be derived from Theorem 3 of \citet{COLT:2022:Garber}
\begin{corollary}
    \label{corollary:static_regret}
      Under Assumptions \ref{assump:K-bound}, \ref{assump:Lipschitz} \ref{assump:convex},  Algorithm \ref{alg2} with the same parameter setting in Theorem \ref{theorem:dynamic_BOGD} guarantees a static regret  bound of 
    \begin{equation}
         \text{{\rm Regret}}_{T}  \leq \O(T^{3/4}).
    \end{equation}
\end{corollary}

\subsection{Improved projection-free dynamic regret}
Note that the linear dependency on $P_T$ in Theorem \ref{theorem:dynamic_BOGD} is too loose and the obtained bound can be vacuous with $P_T = \Omega(T^{1/4})$. To address this issue, we propose a two-layer  method, termed as \underline{P}rojection-free  \underline{O}nline  \underline{L}earning with \underline{D}ynamic Regret (POLD), with a tighter  bound of  $\O(T^{3/4}(1+P_T)^{1/4})$. To help understanding, we first briefly introduce the motivation behind POLD.

Let us consider a given sequence $\Tilde{\bu}_1, \cdots, \Tilde{\bu}_T \in \K$ with the path-length $\Tilde{P}_T = \sum_{t=2}^T \|\Tilde{\bu}_{t-1} - \Tilde{\bu}_t\|_2$. According to  Theorem \ref{theorem:dynamic_BOGD}, we can choose the step size $\Tilde{\eta} = \O(T^{-3/4}(1+\Tilde{P}_T)^{3/4})$ and achieve a tighter $\O(T^{3/4}(1+\Tilde{P}_T)^{1/4})$ bound.
This indicates that if the path-length is known, we can actually tune the step size to obtain an improved bound. 
To deal with the uncertainty of the path-length, we adopt the strategy of maintaining multiple step sizes \citep{NeurIPS:2016:Erven,NeurIPS:2018:Zhang}, and leverage  the two-layer structure: running multiple $\text{BOGD}_\text{IP}$ algorithms with different step sizes and combining them by a meta algorithm.
In the following, we describe the detailed procedure.

First, we create a set of step sizes 
\begin{equation}
    \label{eq:POLD_set}
    \mathcal{H} = \left\{ \left.  \eta_i =  2^{i-1} \left(\frac{7D^2}{ 2 G^2 T}\right)^{3/4} ~\right|~i=1, \cdots, N \right\}, 
\end{equation}
where $N = \lceil \frac{3}{4}\log_2 (1+4T/7)\rceil + 1$. Then, we  activate a set of experts $\{E_i  \mid  \eta_i \in \mathcal{H}\}$, each of which is an instance of $\text{BOGD}_\text{IP}$  with the step size $\eta_i$ chosen from  $\H$.
For each expert $E_i$, we initiate its weight $w_1^i = \frac{C}{i(i+1)}$ where $C = 1 + \frac{1}{N}$. 
Next, inspired by the Hedge algorithm \citep{Others:1997:Freund}, we combine the actions of experts in a weighted-average fashion. Concretely, in each round $t$, POLD receives the action $\bx_t^i$ from expert $E_i$, and computes the weighted average action: 
\begin{equation}
    \label{eq:APOL-x}
    \bx_t = \sum_{i \in \mathcal{H}} w_t^i \bx_t^i, 
\end{equation}
where $w_t^i$ is the weight assigned to $E_i$.
After that, POLD updates the weight of $E_i$ by 
\begin{equation}
    \label{eq:APOL-w}
    w_{t+1}^i = \frac{w_t^i e^{-\alpha f_{t}(\bx_t^i)}}{\sum_{\mu \in \mathcal{H}} w_t^\mu e^{-\alpha f_{t}(\bx_t^\mu)}}, 
\end{equation}
where $\alpha$ denotes  the learning rate of the meta algorithm. Finally, POLD reveals the function $f_t(\cdot)$ to all experts so that they can update their actions for the next round. We summarize all the procedure in Algorithm \ref{alg3}, and present the following theorem.

\begin{algorithm}[tb]
    \caption{\underline{P}rojection-free  \underline{O}nline   \underline{L}earning with \underline{D}ynamic Regret (POLD)}
    \label{alg3}
    \textbf{Input}: A learning rate $\alpha$, a set $\mathcal{H}$ containing step size $\eta_i$ for each expert $E_i$

    \textbf{Initialization}: Activate a set of experts $\{E_i  \mid  \eta_i \in \mathcal{H}\}$ by invoking $\text{BOGD}_{\text{IP}}$ for each $\eta_i \in \mathcal{H}$.

    \begin{algorithmic}[1]
        \STATE For each expert $E_i$, set $w_1^i = \frac{C}{i(i+1)}$ where $C = 1 + \frac{1}{N}$
        
        \FOR{$t=1$ to $T$}
        \STATE Receive $\bx_t^i$ from each expert $E_i$
	    	
        \STATE Compute $\bx_t$ according to \eqref{eq:APOL-x}

        \STATE Submit $\bx_t$, and update the weight $w_{t+1}^i$ for each expert $E_i$ according to \eqref{eq:APOL-w}

        \STATE Send $f_t(\cdot)$ to each expert $E_i$

        \ENDFOR

    \end{algorithmic}
\end{algorithm}

\begin{theorem}
    \label{theorem:Ader_BOGD}
    Let $\alpha = \sqrt{8/T}$ and $\mathcal{H}$ be defined as \eqref{eq:POLD_set}. Under Assumptions \ref{assump:K-bound}, \ref{assump:Lipschitz}, \ref{assump:convex}  and \ref{assump:f-bound},  Algorithm \ref{alg3} guarantees
    \begin{equation*}
        \begin{split}
            \text{{\rm D-Regret}}_T(\bu_1, \cdots, \bu_T)  &\leq \mathcal{O}\left(T^{3/4}(1+P_T)^{1/4}\right).
        \end{split}
    \end{equation*}
\end{theorem}
\textbf{Remark:} Compared with the upper bound in Theorem \ref{theorem:dynamic_BOGD}, the dependence on the path-length is reduced from $P_T$ to $ P_T^{1/4}$.

\subsection{Projection-free adaptive regret}
As mentioned before, besides the dynamic regret \eqref{eq:general_dynamic_regret}, there do exist another metric called (strongly) adaptive regret \eqref{eq:strong_adaptive_regret} in the non-stationary environments. In this section, we  proceed to investigate minimizing \eqref{eq:strong_adaptive_regret}  and present \underline{P}rojection-free  \underline{O}nline  \underline{L}earning with \underline{A}daptive Regret (POLA). Following existing studies on adaptive regret \citep{ECCC:2007:Hazan,ICML:2015:Daniely}, POLA contains  three parts: an expert algorithm, a set of intervals, and a meta algorithm. In the following, we specify them separately. 

\begin{algorithm}[tb]
    \caption{\underline{P}rojection-free  \underline{O}nline  \underline{L}earning with \underline{A}daptive Regret (POLA)}
    \label{alg-APOD}

    \begin{algorithmic}[1]
		
        \FOR{$t=1$ to $T$}
        \FOR{$I \in \mathcal{C}_t$}
        \STATE Create an expert $E_I$ which runs $\text{BOGD}_{\text{IP}}$ from an arbitrary initial point with  $\eta = |I|^{-3/4}$
        
        \STATE For the expert $E_I$, set $R_{t-1, I} = C_{t-1, I} = 0$
    
        \STATE Add expert $E_I$ to the set of active experts $\mathcal{A}_t$
        
        \ENDFOR
        
        \STATE From $\mathcal{A}_t$, remove all experts who end at the round $t$
    
        \STATE Receive the action $\bx_{t, I}$ of each expert $E_I \in \mathcal{A}_t$ and calculate its weight $w_{t, I}$ according to \eqref{eq:APOD_p_t_I}

        \STATE Submit  $\bx_{t}$  defined in \eqref{eq:APOD_x_t} and then receive $f_t(\cdot)$

        \STATE For each $E_I \in \mathcal{A}_t$, update
        \begin{equation*}
            \begin{split}
                R_{t,I} &= R_{t-1, I} + f_t(\bx_t) - f_t (\bx_{t, I})\\
                C_{t,I} &= C_{t-1, I} + |f_t(\bx_t) - f_t (\bx_{t, I})|
            \end{split}
        \end{equation*}

        \STATE Send $f_t(\cdot)$ to each expert $E_I \in \mathcal{A}_t$

        \ENDFOR

    \end{algorithmic}
\end{algorithm}

\begin{figure*}[t]
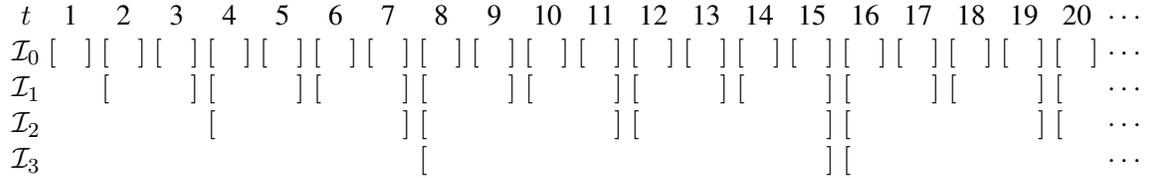

    \setlength\tabcolsep{1.5pt}
    \begin{center}
    \begin{tabular}{ccccccccccccccccccccccccccccccccc}
        $t$  &1 &2 &3 &4 &5 &6 &7 &8 &9 & 10 & 11 &12 &13 &14 &15 &16 &17 &18 &19 &20 & $\cdots$\\
         ${\I}_0$  & $[\quad]$ & $[\quad]$ & $[\quad]$ & $[\quad]$ & $[\quad]$ & $[\quad]$ & $[\quad]$ & $[\quad]$ & $[\quad]$ & $[\quad]$ & $[\quad]$ & $[\quad]$ & $[\quad]$ & $[\quad]$ & $[\quad]$ & $[\quad]$ & $[\quad]$ & $[\quad]$ & $[\quad]$ & $[\quad]$ & $\cdots$\\
         ${\I}_1$  & \text{ } & $[\quad$\text{ }&\text{ }$\quad]$ & $[\quad$\text{ }&\text{ }$\quad]$ & $[\quad$\text{ }&\text{ }$\quad]$ & $[\quad$\text{ }&\text{ }$\quad]$ & $[\quad$\text{ }&\text{ }$\quad]$ & $[\quad$\text{ }&\text{ }$\quad]$ & $[\quad$\text{ }&\text{ }$\quad]$ & $[\quad$\text{ }&\text{ }$\quad]$ & $[\quad$\text{ }&\text{ }$\quad]$ & $[\quad$\text{ }&$\cdots$\\
         ${\I}_2$ & \text{ } & \text{ } & \text{ }& $[\quad$\text{ }&\text{ }&\text{ }&\text{ }$\quad]$ &
         $[\quad$\text{ }&\text{ }&\text{ }&\text{ }$\quad]$ &
         $[\quad$\text{ }&\text{ }&\text{ }&\text{ }$\quad]$ &
         $[\quad$\text{ }&\text{ }&\text{ }&\text{ }$\quad]$ &
         $[\quad$\text{ }&$\cdots$\\
         ${\I}_3$ 
         &\text{ }&\text{ }&\text{ }&\text{ }& \text{ }&\text{ }&\text{ }& $[\quad$\text{ }&\text{ }&\text{ }&\text{ }& \text{ }&\text{ }&\text{ }&\text{ }$\quad]$ 
         & $[\quad$\text{ }&\text{ }&\text{ }&\text{ }&\text{ }&$\cdots$\\
    \end{tabular}
    \caption{Geometric Covering (GC) intervals. In the figure, each interval is denoted by $[\quad]$.}
    \label{fig:GC}
    \end{center}
    \vspace{-16pt}
\end{figure*}

First, we take $\text{BOGD}_\text{IP}$ as the expert algorithm, since it is projection-free and ensures an $\O(|I|^{3/4})$ static regret for a given interval $I$ as shown in Corollary \ref{corollary:static_regret}.
Then, we build the GC intervals \citep{ICML:2015:Daniely} shown in Figure~\ref{fig:GC}: 
\begin{equation}
     \I= \bigcup_{k \in \mathbb{N} \cup \{0\}} \I_k,\quad \I_k=\left\{ [ i \cdot 2^k, (i+1) \cdot 2^k -1]: i \in \mathbb{N} \right\}.
\end{equation}
For each interval $I$, we maintain an instance of $\text{BOGD}_\text{IP}$, denoted as the expert $E_I$, to minimize the static regret over that interval. According to  Corollary \ref{corollary:static_regret}, we set the step size $\eta = |I|^{-3/4}$ to  obtain the $\O(|I|^{3/4})$ static regret bound over the interval $I$.

Next, to track the best expert on the fly, we choose AdaNormalHedge \citep{COLT:2015:Luo} as the meta algorithm since it naturally supports the setting that the number of experts varies over time \citep{ICML:2019:Zhang}. The key ingredient of AdaNormalHedge is the potential function:
$
    \Phi(R, C) = \exp{ ( [R]_+^2/3C )}, 
$
where $[x]_+ = \max(0, x)$, $\Phi(0, 0) = 1$ and $R$, $C$ are two variables maintained by each expert. Based on $\Phi(R, C)$, we can compute the weight for each expert according to the following  weight function:
\begin{equation*}
    w(R,C) = \frac{1}{2}\left( \Phi(R+1, C+1)  -  \Phi(R+1, C-1) \right).
\end{equation*}

Putting all pieces together, we obtain our projection-free POLA for adaptive regret minimization. Below, we describe the detailed procedure, which is also summarized in Algorithm \ref{alg-APOD}. 

For brevity, we denote the set of all active experts as $\A_t$ for the round $t$, and the set of intervals that start from the round $t$ as $\mathcal{C}_{t}=\{ I \mid I \in \I, t \in I,(t-1) \notin I\}$.
In Step 3, we create an instance of $\text{BOGD}_\text{IP}$ as the expert $E_I$ for each $I \in \C_t$, and initiate it from an arbitrary initial point with the step size $\eta = |I|^{-3/4}$. In Step 4, we set the variables $R_{t-1, I} = C_{t-1, I} = 0$ for $E_I$, where $R_{t-1,I} = \sum_{u = \min I}^{t-1} f_t(\bx_t) - f_t (\bx_{t, I})$ denotes the regret of $E_I$ up to round $t-1$, and 
$C_{t-1,I} = \sum_{u = \min I}^{t-1} |f_t(\bx_t) - f_t (\bx_{t, I})|$ denotes the   sum of the absolute value of instantaneous regrets, 
and $\min I$ denotes the beginning round of $I$. 
In Step 5, the new expert $E_I$ is added to $\A_t$. Then, we remove all experts from $\A_t$, who end at the round $t$ (Step 7).
After receiving the action $\bx_{t,I}$ from $E_I$, we update its corresponding weight as following:
\begin{equation}
    \label{eq:APOD_p_t_I}
    w_{t, I} = \frac{w(R_{t-1,I}, C_{t-1,I})}{\sum_{E_I \in \A_t} w(R_{t-1,I}, C_{t-1,I})}.
\end{equation}
In Step 9, we submit the weighted action 
\begin{equation}
    \label{eq:APOD_x_t}
    \bx_t = \sum_{E_I \in \A_t} w_{t,I} \bx_{t, I}, 
\end{equation}
and receive the loss function $f_t(\cdot)$. In Step 10, for each $E_I \in \A_t$, we compute its corresponding variables $R_{t,I}$ and $ C_{t,I}$. At the end, we reveal $f_t(\cdot)$ to all active experts, so that they can update their actions for the next round (Step 11).
We present the adaptive regret bound of POLA below.
\begin{theorem}
    \label{theorem:APOD-adaptive_regret}
    Under Assumptions \ref{assump:K-bound}, \ref{assump:Lipschitz}, \ref{assump:convex} and \ref{assump:f-bound}, Algorithm \ref{alg-APOD} guarantees
    \begin{equation*}
        \text{{\rm SA-Regret}}_T(\tau) \leq \O(\sqrt{\tau \log T} + \tau^{3/4}) = \Tilde{\O}\left(\tau^{3/4}\right).
    \end{equation*}
\end{theorem}
\textbf{Remark:} Compared to  existing methods \citep{COLT:2022:Garber,ALT:2023:Lu} for adaptive regret minimization, POLA has following advantages.

\begin{compactitem} [\hspace{0.00cm}$\bullet$]
     \item POLA enjoys an $\Tilde{\O}(\tau^{3/4})$ strongly adaptive regret, and thus can still perform well on  short intervals. In contrast, \citet{COLT:2022:Garber} minimize the weak adaptive regret \eqref{eq:weak_adaptive_regret}, which only promises a performance guarantee on long intervals. 

    \item For each expert, POLA performs only $\O(1)$ linear optimizations per round on average, whereas \citet{ALT:2023:Lu} require a significantly higher  number of $\O(\log T)$ membership operations.     
    Moreover, their operations could be inefficient compared to linear optimizations in many popular domains. For example, the trace norm constraints $\K = \{ X |  \|X\|_* \leq \delta, X \subset \mathbb{R}^{m \times n}\}$ incurs a  membership operations cost of $\O(mn^2)$ while the linear optimization cost is $\O(nnz(X))$, where $nnz(X)$ denotes the number of non-zero entries \citep{COLT:2022:Mhammedi}.
\end{compactitem}  

Moreover, we note that previous studies on projection-based online learning \citep{AISTATS:2020:Zhang,ICML:2020:Cutkosky} have shown that it is possible to design a single algorithm to minimize dynamic regret and adaptive regret simultaneously. In particular, our POLA shares a similar two-layer structure with the method of \citet{AISTATS:2020:Zhang}, inspiring us to investigate the performance of POLA for dynamic regret minimization. The following theorem shows that POLA also enjoys an $\O(T^{3/4}(1+P_T)^{1/4})$ dynamic regret bound.
\begin{theorem}
    \label{theorem:APOD-dynamic_regret}
    Under Assumptions \ref{assump:K-bound}, \ref{assump:Lipschitz}, \ref{assump:convex} and \ref{assump:f-bound}, Algorithm \ref{alg-APOD} guarantees
    \begin{equation*}
        \text{{\rm D-Regret}}_T(\bu_1, \cdots, \bu_T) \leq \mathcal{O}\left(T^{3/4}(1+P_T)^{1/4}\right).
    \end{equation*}
\end{theorem}
\textbf{Remark:} Although POLA achieves the same dynamic regret bound as POLD,  this does not imply that the latter one is insignificant. Compared with POLA, POLD employs a  simpler meta algorithm and does not need to construct  GC intervals,  making it  much easier to comprehend and implement.

\section{Experiments}
\label{sec:exp}

In this section, we present numerical experiments to support our theoretical results in dynamic regret and adaptive regret minimization. All experiments are conducted on a machine equipped with the Intel Xeon E5-2620 CPU and 32G memory, and each of them is repeated five times with different random seeds. We present experimental results (mean and standard deviation) in Figures \ref{fig:1} and  \ref{fig:2}.

\subsection{Dynamic regret minimization}

\begin{figure*}[t] 
    \begin{minipage}[t]{0.33\linewidth}
        \centering
        \includegraphics[width=2.0in]{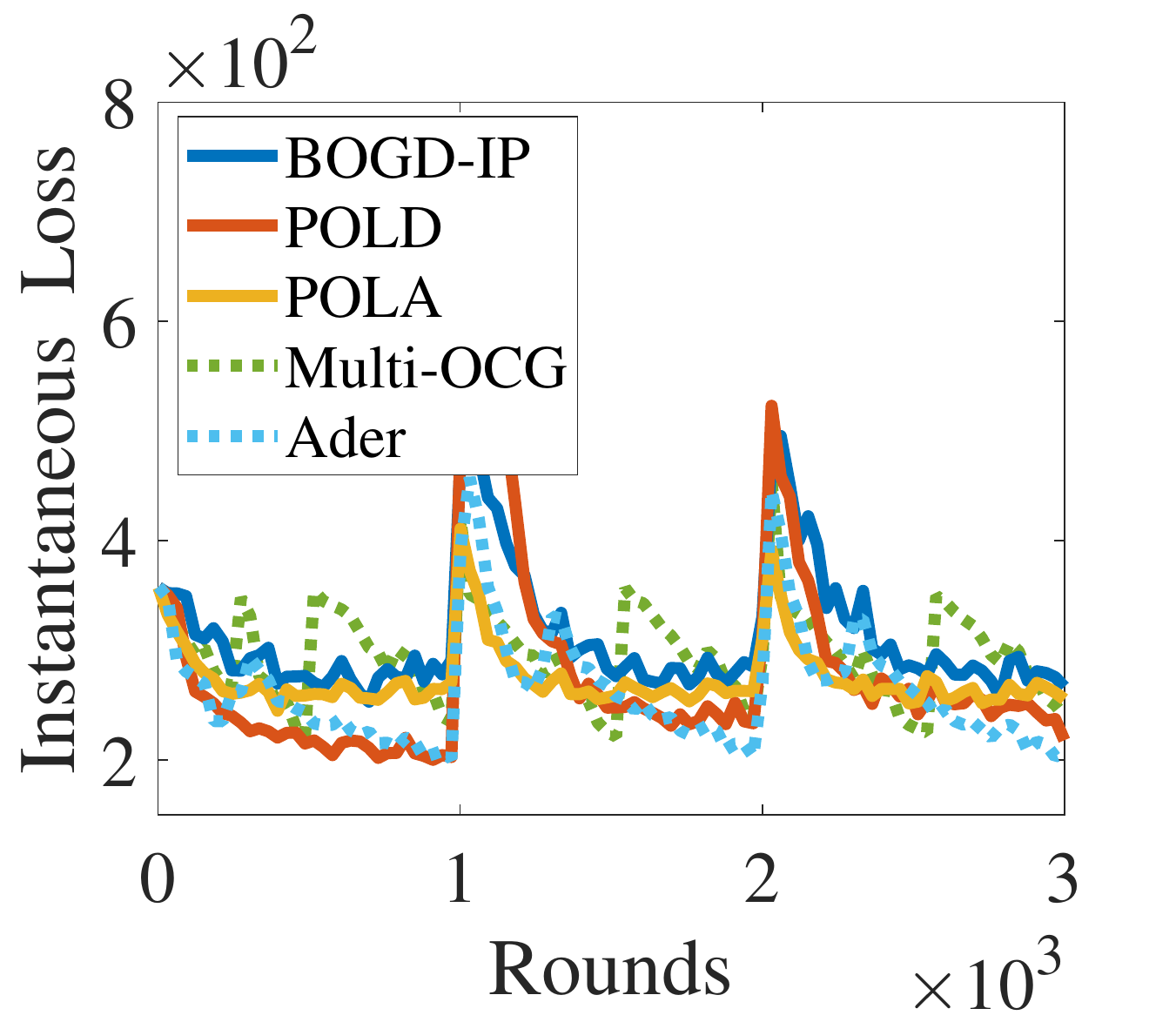} 
    \end{minipage}
    \begin{minipage}[t]{0.33\linewidth}
        \centering
        \includegraphics[width=2.0in]{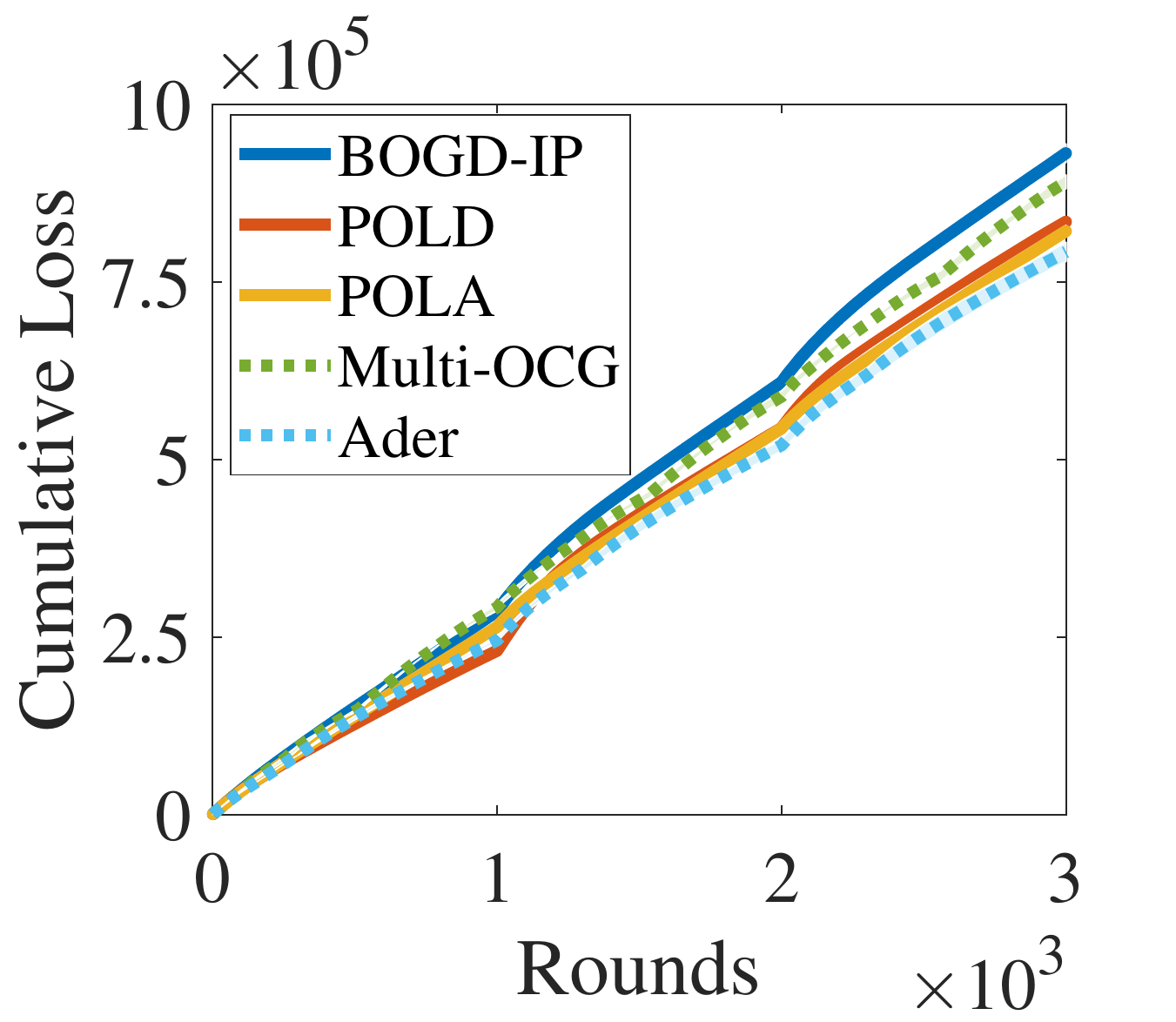}
    \end{minipage}%
    \begin{minipage}[t]{0.33\linewidth}
        \centering
        \includegraphics[width=2.0in]{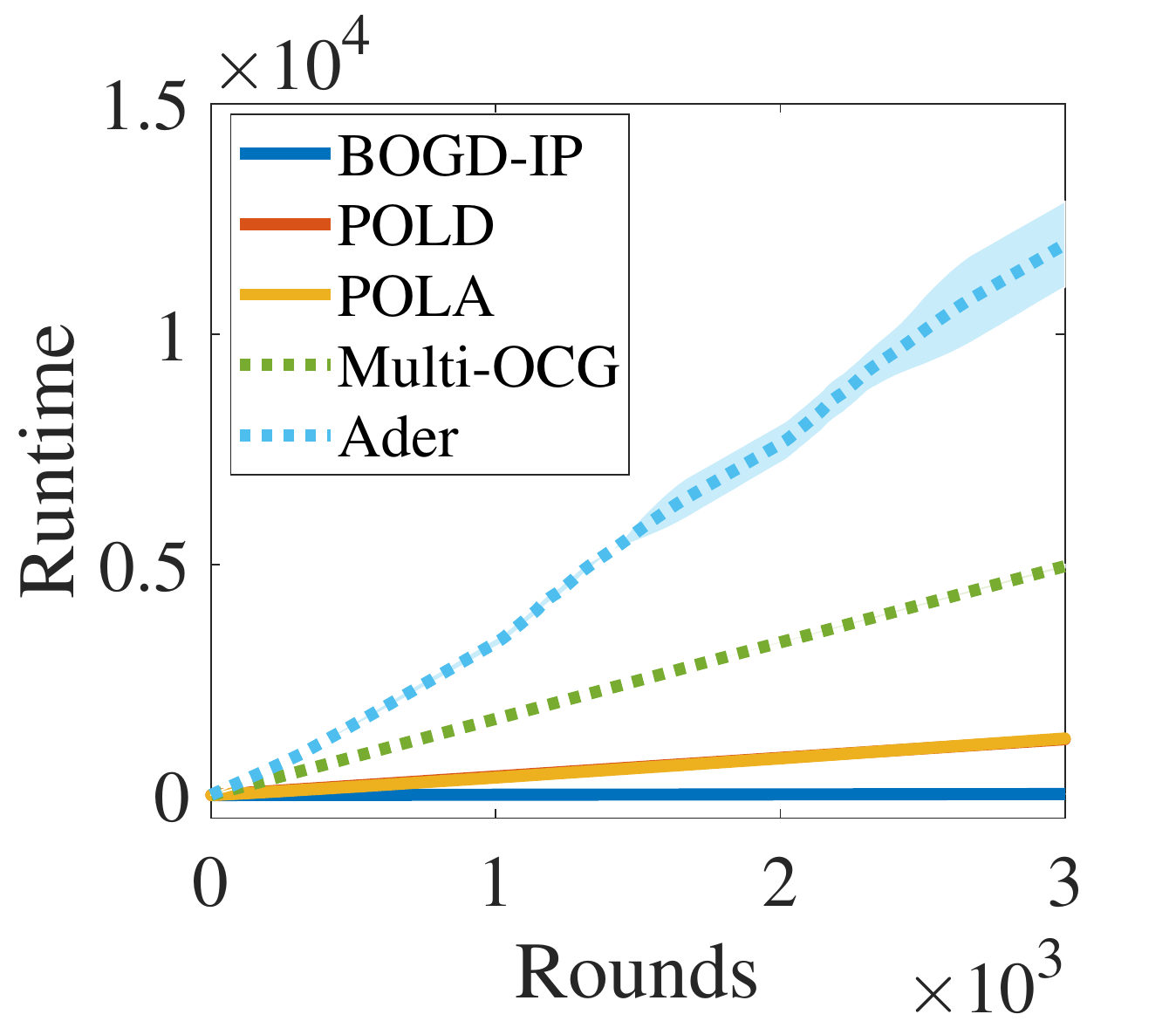}
    \end{minipage}
    \caption{Experimental results  for dynamic regret minimization.}
    \label{fig:1}
\end{figure*}

\paragraph{Setup.} To evaluate our methods (i.e. $\text{BOGD}_\text{IP}$, POLD and POLA) in dynamic regret minimization, we study the problem of online matrix completion, of which the goal is to produce a matrix $X$ from the trace norm ball in an online fashion to  approximate the target matrix $M \in \mathbb{R}^{m \times n}$.  Specifically, in each round $t$, the learner receive a sampled data $(i, j)$ with the value $M_{ij}$ from the entry set $OB$ of $M$. 
Then, the learner chooses  $X$ from the trace norm ball $\K = \{ X |  \|X\|_* \leq \delta, X \subset \mathbb{R}^{m \times n}\}$  where $\delta$ is the parameter, and suffers the online  loss 
\begin{equation*}
    f_t(X) = \sum_{(i, j) \in OB} | X_{ij} - M_{ij} |.
\end{equation*}
We conduct the experiments with $\delta=10^4$ on the public dataset: MovieLens 100K\footnote{https://grouplens.org/datasets/movielens/100k/}, which contains 100000 ratings from 943 users on 1682 movies. Following \citet{AAAI:2021:Wan:B}, we slightly modify the dataset to simulate the non-stationary environments. Concretely, we generate an extended  datasets $\{(i_k, j_k, M_{i_kj_k})\}_{k=1}^{300000}$ by merging three copies of MovieLens $100$K. For entries corresponding to $k = 100001, \cdots, 200000$, we negate the original values $M_{i_kj_k}$ to obtain $-M_{i_kj_k}$. For simplicity, we divide the extended datasets into $T=3000$ partitions. In this way, the target matrix $M$ drifts every $1000$ rounds.

\paragraph{Contenders.} We compare  our methods with the projection-free algorithm: Multi-OCG \citep{AAAI:2021:Wan:B}, and the projection-based algorithm: Ader \citep{ICML:2018:Zhang}. All parameters of each method are set according to the theoretical suggestions. For instance, the learning rate of the $i$-th expert is set as $\eta_i = c \left(2^{i-1}\right)^{-1/2}$ in Multi-OCG, and $\eta_i = c  2^{i-1}   T^{-1/2}$ in Ader, and $\eta_i = c  2^{i-1} T^{-3/4}$ in POLD, where $c$ is the hyper-parameters selected   from $\{10^{-1}, 10^{0}, \cdots, 10^{6}\}$.

\paragraph{Results.} We report  the average instantaneous  loss, the cumulative loss and the runtime (in seconds) against the number of rounds for each method in Figure \ref{fig:1}.  As evident from the results, projection-free methods are significantly more efficient compared to the projection-based  approach (i.e. Ader), albeit with a slight compromise on cumulative loss. This observation is reasonable in the sense that (i) the cost of linear optimization over the trace norm ball is $\O(nnz(X))$ whereas projection operation  suffers a much higher $\O(mn^2)$  cost; (ii) our methods ensure an $\O(T^{3/4}(1+P_T)^{1/4})$ bound against the $\O(\sqrt{T(1+P_T)})$ bound of Ader. Moreover, owing to the inherent advantage in minimizing the general-case dynamic regret, our methods yield a lower cumulative  loss compared to  the projection-free contender Multi-OCG.

\subsection{Adaptive regret minimization}
\begin{figure*}[t] 
    \begin{minipage}[t]{0.33\linewidth}
        \centering
        \includegraphics[width=2.0in]{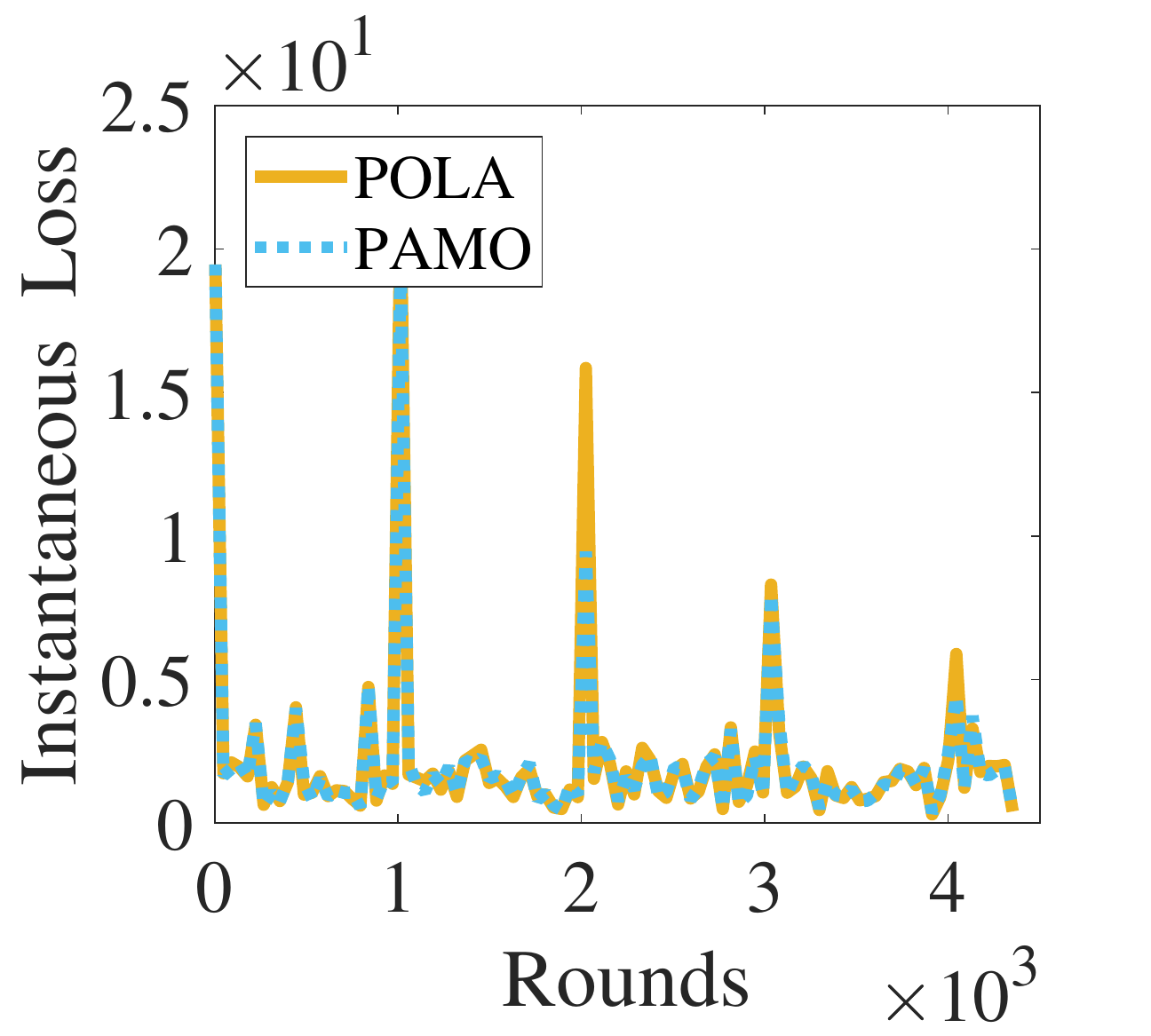} 
    \end{minipage}
    \begin{minipage}[t]{0.33\linewidth}
        \centering
        \includegraphics[width=2.0in]{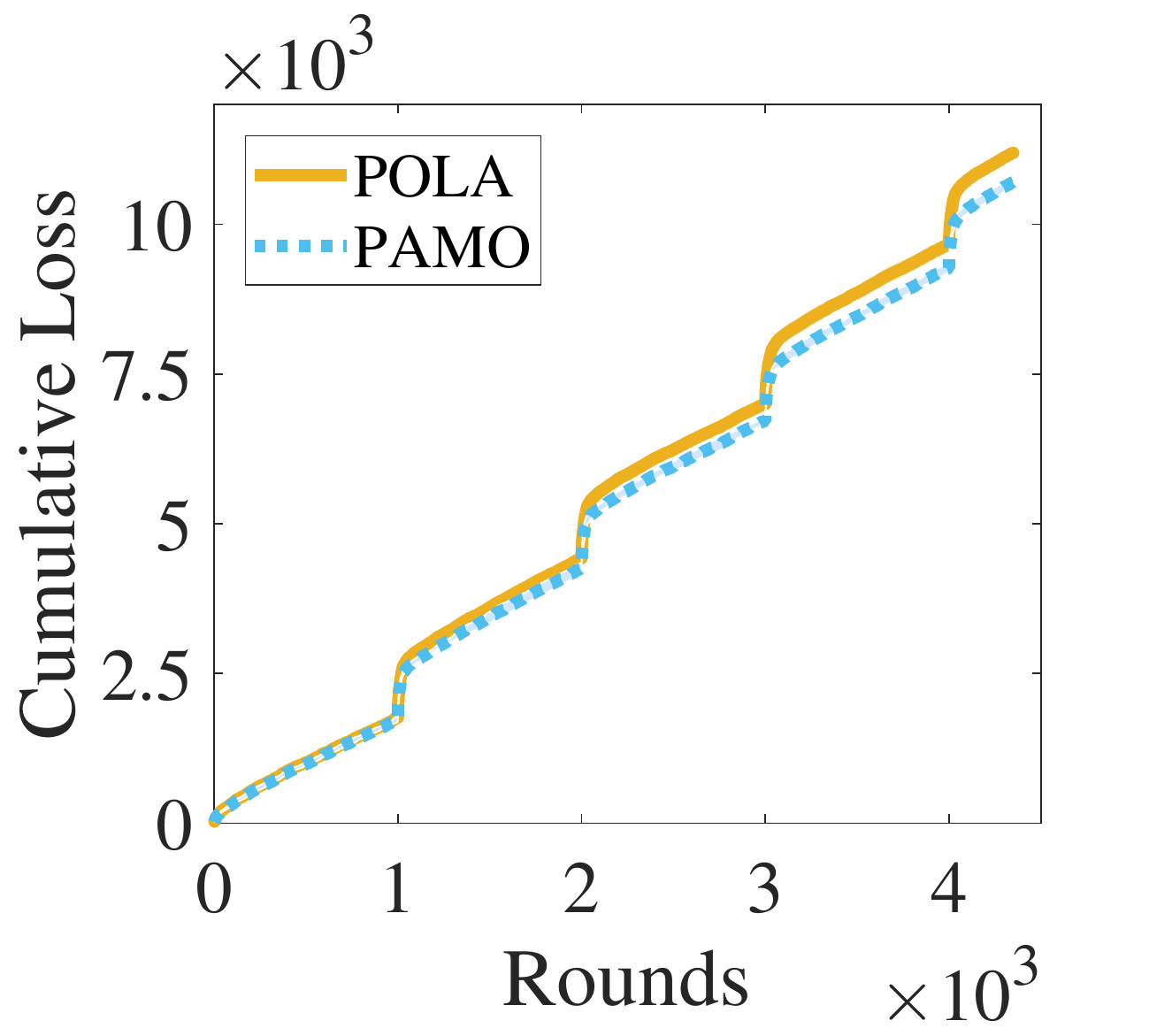}
    \end{minipage}%
    \begin{minipage}[t]{0.33\linewidth}
        \centering
        \includegraphics[width=2.0in]{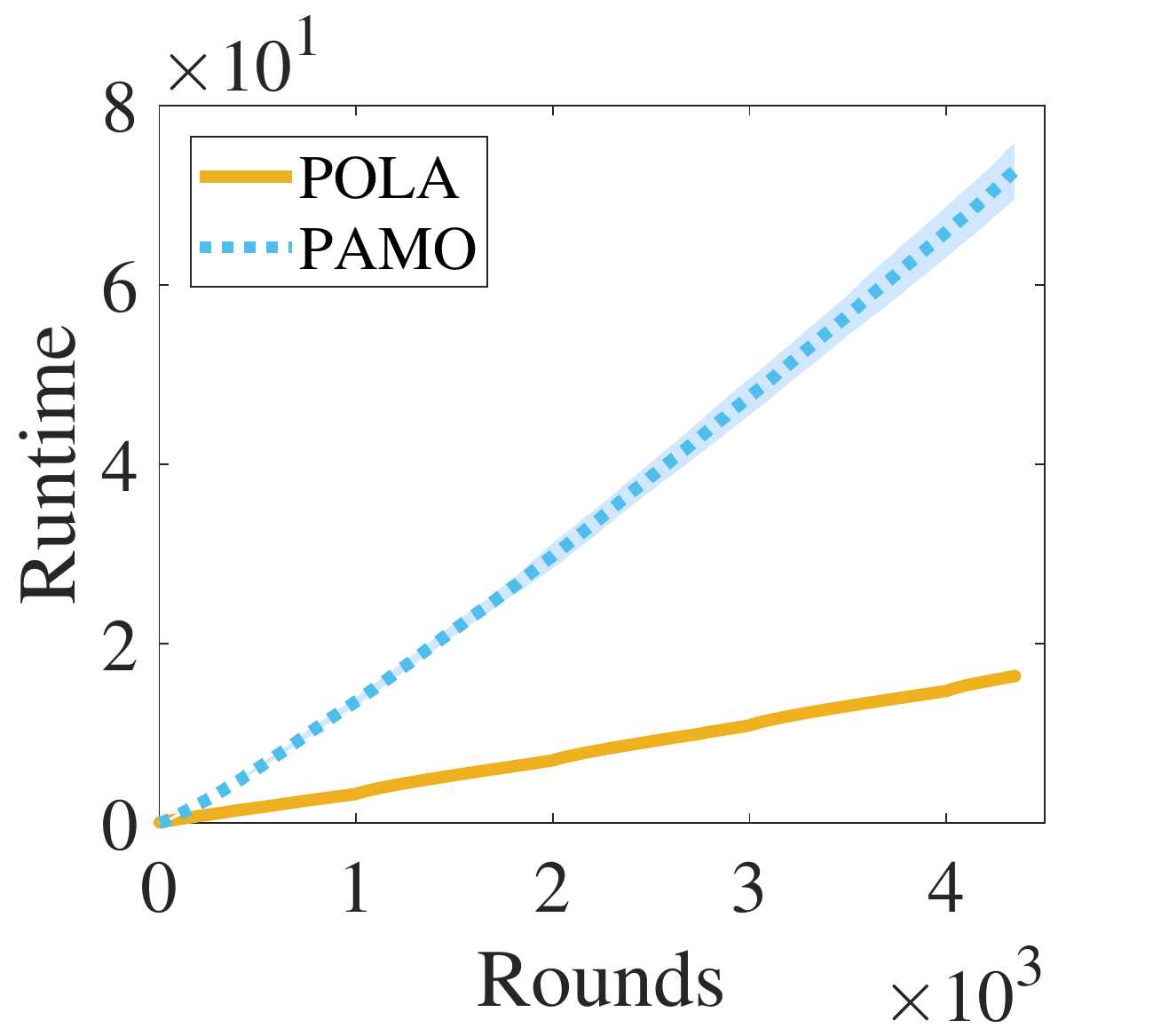}
    \end{minipage}
    \caption{Experimental results  for adaptive regret minimization.}
    \label{fig:2}
\end{figure*}

\paragraph{Setup.} To evaluate our method (i.e., POLA)  in adaptive regret minimization, we consider the problem of online multiclass  classification. In each round $t$, the learner is presented a sampled data $(\be_t, l_t)$ with $\be_t\in \mathbb{R}^d$ being the feature and $l_t \in \mathcal{C} = \{1, \cdots, h\}$ being the corresponding class label. Then, the learner is required to choose a decision matrix $X = [\bx_1, \cdots, \bx_h]^\top \in \mathcal{K} $, where  $\K = \{ X |  \|X\|_* \leq \delta, X \subset \mathbb{R}^{h \times d}\}$ denotes the trace norm ball with the parameter $\delta$, and predict the class label as $\argmax_{j \in \mathcal{C}}  \bx_j^\top \be_t$. Next, the learner incurs a convex multivariate logistic loss 
\begin{equation*}
    f_t(X) =  \log \left( 1 + \sum_{j \neq l_{t}} \exp{\left( \bx_{j}^{\top}  \be_{t} - \bx_{l_{t}}^{\top}  \be_{t}\right)} \right).
\end{equation*}
We perform experiments with $\delta=10^2$ on the public shuttle dataset \citep{Others:2011:Chang}, which contains $43500$ data belonging to $7$ classes. For bevity, the dataset is divided equally into $4350$ partitions, and  we flip the original features  by multiplying $-1$ every $1000$ partitions to simulate the non-stationary environments.

\paragraph{Contenders.} To verify the performance
benefit of POLA by solving linear optimizations, we choose \underline{P}rojection-free \underline{A}daptive via a \underline{M}embership \underline{O}racle (PAMO) \citep{ALT:2023:Lu}, which employs membership oracle in lieu of the projection operation, as the contender.  All parameters of each method are set according to the theoretical suggestions. For instance, the learning rate of the $i$-th expert is set as $\eta_i = c \left(2^{i-1}\right)^{-1/2}$ in PAMO and $\eta_i = c  \left(2^{i-1}\right)^{-3/4}$ in POLA, where $c$ is the hyper-parameters selected   from $\{2^{-4}, 2^{-3}, \cdots, 2^{4}\}$.

\paragraph{Results.}

We plot the average instantaneous  loss, the cumulative loss and the runtime  (in seconds) of each algorithm in Figure \ref{fig:2}.  It can be clearly seen that POLA is significantly faster than PAMO, despite a slight  sacrifice of cumulative loss. This is reasonable since each invocation of membership oracle over the trace norm ball $\K$ requires $\O(h^2 d)$ costs, while linear oracle suffers only $\O(nnz(X))$. Therefore, as mentioned previously, each expert in PAMO suffers a total of $\O(h^2 d \log T)$ computational costs per round, compared to $\O(nnz(X))$ in POLA.

\section{Conclusion and future work}

In this paper, we investigate non-stationary projection-free online learning with dynamic regret and adaptive regret guarantees. Specifically, in the dynamic regret minimization, we provide a novel dynamic regret analysis for  $\text{BOGD}_\text{IP}$ \citep{COLT:2022:Garber}, and establish the first $\O(T^{3/4}(1+P_T))$ general-case dynamic regret. Then, we improve this bound to $\O(T^{3/4}(1+P_T)^{1/4})$  by proposing POLD, which runs a set of $\text{BOGD}_\text{IP}$ algorithms with different step sizes in parallel  and tracks the best one on the fly. In the adaptive regret minimization, we present our  method POLA with an $\Tilde{\O}(\tau^{3/4})$  strongly adaptive regret bound.  The essential idea is to construct the GC intervals, maintain an instance of $\text{BOGD}_\text{IP}$ to minimize the static regret for each interval, and then combine actions of instances by a meta algorithm. Furthermore, we show that POLA can also minimize the dynamic regret and achieve the same bound as that of POLD. Empirical studies on dynamic regret and adaptive regret minimization have verified our theoretical findings.

Currently, both POLD and POLA need to maintain $\O(\log T)$ experts, which leads to $\O(\log T)$ linear optimizations per round. Therefore, a natural question arises:  is it possible to further reduce the number of linear optimizations in each round, i.e.,~from $\O(\log T)$ to $\O(1)$? We note that in non-stationary projection-based online learning, $\O(\log T)$ projection operations can indeed be reduced to $\O(1)$ \citep{NeurIPS:2022:Zhao}. But in the projection-free setting, it seems highly non-trivial and we leave it as a future work.

\bibliography{ref}
\bibliographystyle{plain}

\appendix
\newpage

\section{Infeasible projection oracle}
\label{sec:IP}
For the sake of completeness, we provide a brief description of infeasible projection oracle $\O_{\text{IP}}$,  which can also be found in \citet{COLT:2022:Garber}.

Different from traditional projection operations, the infeasible projections consider projecting the original point back to an infeasible point, which could be still at the outside of  the domain $\K$, but is sufficiently close to $\K$. \citet{COLT:2022:Garber} show that such a transform is effective, and can be efficiently achieved by calling to the infeasible projection oracle $\O_{\text{IP}}$ over the domain $\K \subseteq R \B$ with the error tolerance $\epsilon$ as follows.
\begin{equation*}
    \bx, \Tilby  =  \mathcal{O}_{\text{IP}}(\mathcal{K}, \epsilon, \bx_0, \by_0).
\end{equation*}
Specifically, given the inputs $(\bx_0, \by_0) \in \K \times \mathbb{R}^d$, the infeasible projection oracle $\O_{\text{IP}}$  returns $(\bx, \Tilby) \in \K \times R \B$, where $\|\bx - \Tilby\|_2 \leq \sqrt{3 \epsilon}$ and  $\forall \bz \in \K, \|\Tilby  - \bz\|_2 \leq \|\by_0  - \bz\|_2$. 
\citet{COLT:2022:Garber} provide an implementation of $\O_{\text{IP}}$ by solving linear optimizations, which repeats following two steps: (i) computing the feasible point $\bx \in \K$ that is  sufficiently close to the infeasible point $\Tilde{\by}$; (ii) ``pulling" $\Tilde{\by}$ close to $\K$ based on the feasible point $\bx \in \K$. In the following, we sketch this procedure.

First, we briefly introduce how to obtain the action $\bx \in \K$, which is also summarized in Algorithm \ref{alg:FW-SH}. We compute the feasible action $\bx$ from the initial point $\bx_1$ iteratively. This procedure can be viewed as a variant of  Frank-Wolfe algorithm with line-search, and the loss function is $f(\bx)=\frac{1}{2}\|\bx - \by\|_2^2$ (Step 2 and Step 6-7). The main different lies in Step 3-5, which indicates the stop condition and takes responsibility for iterations. We can prove that the output $\bx \in \K$ ensures following theoretical guarantees.
\begin{lemma}
    \label{lemma:FW-iterations}
    (Lemma 6 in \citet{COLT:2022:Garber}) Under Assumption \ref{assump:K-bound}, for a fixed error tolerance $\epsilon$, Algorithm \ref{alg:FW-SH} stops after at most $\lceil 27D^2/4\epsilon - 2\rceil$ iterations, and the output $\bx$ guarantees:
    \begin{itemize}
        \item $\|\bx - \by\|_2^2 \leq \|\bx_1 - \by\|_2^2$.

        \item $\|\bx - \by\|_2^2 \leq 3\epsilon$ ~ or ~ $\forall \bz \in \mathcal{K}: (\by - \bz)^\top (\by - \bx) > 2\epsilon$.

        \item If $\text{dist}^2(\by, \mathcal{K}) < \epsilon$ then $\|\bx - \by\|_2^2 \leq 3\epsilon$
    \end{itemize}
\end{lemma}

Then, we proceed to compute an infeasible point $\Tilde{\by}$ after obtained $\bx \in \K$ (Algorithm \ref{alg:IP}). 
Specifically, we compute $\Tilde{\by}$ iteratively, and the loop stops until $\|\bx - \Tilde{\by}\|_2^2 \leq 3\epsilon$.
In Step 8, the  intermediate point $\by$ is updated along the direction $\by - \bx$ with the step size $\gamma$. Notably, we can prove that the output $\Tilde{\by}$ gets closer to $\K$ during this iteration, which is summarized in following lemma.

\begin{lemma}
    \label{lemma:lemma-IP}
    (Lemma 7 in \citet{COLT:2022:Garber}) Under Assumption \ref{assump:K-bound}, for a fixed error tolerance $\epsilon$ and  $\gamma = \frac{2\epsilon}{\|\bx_0 - \by_0\|_2^2}$,  Algorithm \ref{alg:IP} stops after at most $\max \{\|\bx_0 - \by_0\|_2^2 (\|\bx_0 - \by_0\|_2^2 - \epsilon) / 4 \epsilon^2 + 1, 1 \}$ iterations, and returns $(\bx, \Tilde{\by}) \in \mathcal{K} \times R\mathcal{B} ~(D = 2R)$ which satisfy
    \begin{equation}
        \label{eq:lemma-IP}
        \|\bx - \Tilde{\by}\|_2^2 \leq 3 \epsilon \quad \text{and} \quad \forall \bz \in \mathcal{K}, \|\Tilde{\by} - \bz\|_2^2 \leq \|\by_0 - \bz \|_2^2
    \end{equation}
\end{lemma}

It is noteworthy that the infeasible projection oracle is actually implemented based on the linear optimization over $\K$ (see Step 2 in Algorithm \ref{alg:FW-SH}).

\section{Theoretical analysis}

\begin{algorithm}[tb]
    \caption{Subroutine of  $\mathcal{O}_{\text{IP}}$  (Algorithm 3 in \citet{COLT:2022:Garber})}
    \label{alg:FW-SH}
    \textbf{Input}: Domain set $\mathcal{K}$, error tolerance $\epsilon$, initial point $\bx_1 \in \mathcal{K}$, target point $\by$

    \begin{algorithmic}[1]
        \FOR{$i=1, \cdots $}

        \STATE Compute $\bv_i =  \argmin_{\bx \in \K} \left <\bx_i - \by, \bx\right >$            

        \IF{$(\bx_i - \by)^\top (\bx_i - \bv_i) \leq \epsilon$ or $\|\bx_i - \by\|_2^2 \leq 3 \epsilon$}
        \STATE Return $\bx$ $\leftarrow$ $\bx_i$      

        \ENDIF

        \STATE Set $\delta_i = \argmin_{\delta \in [0,1]}\{\|\bx_i + \delta(\bv_i - \bx_i) - \by\|_2^2  \}$

        \STATE Set $\bx_{i+1} = \bx_i + \delta_i(\bv_i - \bx_i)$

        \ENDFOR

    \end{algorithmic}
\end{algorithm}

\begin{algorithm}[tb]
    \caption{Infeasible Projection Oracle, $\mathcal{O}_{\text{IP}}$ (Algorithm 4 in \citet{COLT:2022:Garber})}
    \label{alg:IP}
    \textbf{Input}: Domain set $\mathcal{K}$,  error tolerance $\epsilon$, initial point $\bx_0 \in \mathcal{K}$, initial point $\by_0$

    \begin{algorithmic}[1]
        \STATE $\by_1 = \by_0 / \max\{1, \|\by\|_2 / R\}$
        \IF{$\|\bx_0 - \by_0\|_2^2 \leq 3 \epsilon$}
        \STATE   Return $(\bx, \Tilde{\by}) \gets (\bx_0, \by_1)$
        \ENDIF
        \FOR{$i=1, \cdots $}

        \STATE $\bx_i \gets $ Algorithm \ref{alg:FW-SH} $(\mathcal{K}, \epsilon, \bx_{i-1}, \by_i)$

        \IF{$ \|\bx_i - \by_i\|_2^2 > 3 \epsilon$}

        \STATE $\by_{i+1} = \by_i - \gamma(\by_i - \bx_i)$

        \ELSE

        \STATE Return $(\bx, \Tilde{\by}) \gets (\bx_i, \by_i)$

        \ENDIF

        \ENDFOR
    \end{algorithmic}
\end{algorithm}

\subsection{Proof of Theorem \ref{theorem:dynamic_BOGD}}
First, we  divide the upper bound of  dynamic regret into three terms.
Let $K$ be the block size, and $t = (m-1)K+k$ be the $k$-th round of the block $m$. By utilizing  the convexity of $f_t(\cdot)$, we have
\begin{equation}
    \label{eq:theorem_2}
    \begin{split}
        &\sum_{t=1}^T f_t(\bx_t) - \sum_{t=1}^T f_t(\bu_t) = \sum_{m=1}^{T/K} \sum_{k=1}^K \left[ f_t(\bx_m)  - f_t(\bu_t)\right]  \overset{\eqref{eq:convex_function}}{\leq}  \underbrace{\sum_{m=1}^{T/K} \sum_{k=1}^K   \left<\nabft(\bx_m), \bx_m - \Tilby_m \right>}_{:=A} \\
        &+ \underbrace{\sum_{m=1}^{T/K} \sum_{k=1}^K  \left<\nabft(\bx_m), \Tilby_m - \bu_{s(m)} \right> }_{:=B}  + \underbrace{\sum_{m=1}^{T/K} \sum_{k=1}^K \left<\nabft(\bx_{m}), \bu_{s(m)} - \bu_{t} \right> }_{:=C},
    \end{split}
\end{equation}
where we assume $T/K$ is an integer  without loss of generality,  and denote $\bu_{s(m)}$ as the comparator of the first round $s(m) = (m-1)K+1$ in block $m$ for brevity. 

Then, we analyze above three terms separately. 
To upper bound term A of \eqref{eq:theorem_2}, we introduce the following lemma.
\begin{lemma}
    \label{lemma:IP}
    Let $\O_{\text{IP}}$ be the infeasible projection oracle over the domain set $\K \subseteq R\B$, and $\epsilon$ be the error tolerance. To compute the action for the block $m$, $\text{BOGD}_\text{IP}$ invokes $\O_{\text{IP}}$ as following
    \begin{equation}
        \label{eq:lemma:IP}
        \bx_{m}, \Tilby_{m} =  \mathcal{O}_{\text{IP}}(\mathcal{K}, \epsilon, \bx_{m-1}, \by_{m}),
    \end{equation}
    and obtains $(\bx_{m}, \Tilby_{m}) \in \mathcal{K} \times R\mathcal{B}$ with 
    \begin{equation}
        \label{eq:lemma:IP_2}
        \|\bx_{m} - \Tilby_{m}\|_2 \leq \sqrt{3\epsilon}.
    \end{equation}
    By choosing proper parameters, each invocation of $\O_{\text{IP}}$ requires solving at most $\O(T^{1/2})$ linear optimizations. 
\end{lemma}
By exploiting Lemma \ref{lemma:IP} and the Lipschitz continuity of $f_t$, the first term $A$ of \eqref{eq:theorem_2} can be upper bounded as follows.
\begin{equation}
    \label{eq:theorem_2_1}
    A =\sum_{m=1}^{T/K} \sum_{k=1}^K   \left<\nabft(\bx_m), \bx_m - \Tilby_m \right> \overset{\eqref{eq:Lipschitz},\eqref{eq:lemma:IP_2}}{\leq} GT \sqrt{3\epsilon}.
\end{equation}

Then, we upper bound the second term $B$ of \eqref{eq:theorem_2}. To this end, we first denote  $P_T'$ as the path-length of the sequence of $\bu_{s(1)}, \cdots, \bu_{s(T/K)}$, i.e.,
\begin{equation}
    P_T' = \sum_{m=2}^{T/K} \|  \bu_{s(m-1)} -  \bu_{s(m)} \|_2,
\end{equation}
and introduce the following lemma.
\begin{lemma}
    \label{lemma:P_T}
    Let $ P_T' = \sum_{m=2}^{T/K} \|  \bu_{s(m-1)} -  \bu_{s(m)} \|_2$ and $ P_T = \sum_{t=2}^{T}  \|  \bu_{t-1} -  \bu_{t} \|_2$, then we have
    \begin{equation}
        P_T' \leq P_T.
    \end{equation}
\end{lemma}
Inspired by the dynamic regret analysis for OGD in \citet{ICML:2003:Zinkevich}, we can obtain the upper bound of $B$ with respect to $P'_T$ as shown below.
\begin{lemma}
    \label{lemma:BOGD_B}
    Let  $ P_T' = \sum_{m=2}^{T/K} \|  \bu_{s(m-1)} -  \bu_{s(m)} \|_2$, and $K$ be the block size and $\eta$ be the step size. Under Assumptions \ref{assump:K-bound} and \ref{assump:Lipschitz}, we have
    \begin{equation}
        \begin{split}
          B  &= \sum_{m=1}^{T/K} \sum_{k=1}^K  \left<\nabft(\bx_m), \Tilby_m - \bu_{s(m)} \right> \leq \frac{7}{4 \eta} D^2 + \frac{\eta}{2}KTG^2 + \frac{D}{\eta}P_T'.
        \end{split}
    \end{equation}
\end{lemma}

By applying Lemma \ref{lemma:P_T} and Lemma  \ref{lemma:BOGD_B}, we have 
\begin{equation}
    \label{eq:theorem_2_5}
    \begin{split}
         B    &= \sum_{m=1}^{T/K} \sum_{k=1}^K  \left<\nabft(\bx_m), \Tilby_m - \bu_{s(m)} \right> \leq \frac{7}{4 \eta} D^2 + \frac{\eta}{2}KTG^2 + \frac{D}{\eta}P_T.
    \end{split}
\end{equation}

Next, we proceed to upper bound the third term $C$ of \eqref{eq:theorem_2}. To simplify the notation, we denote the local path-length during the block $m$ as 
\begin{equation}
    P_T(m) = \sum_{k=2}^{K} \| \bu_{(m-1)K+k-1} - \bu_{ (m-1)K+k}\|_2.
\end{equation}
And for any $ \bu_{(m-1)K+k}~(1 \leq k \leq K)$ in  block $m$, we have
\begin{equation}
    \label{eq:theorem_2_2}
    \begin{split}
         &\| \bu_{s(m)} - \bu_{ t}\|_2 = \| \bu_{s(m)} - \bu_{ (m-1)K+k}\|_2 = \| \bu_{(m-1)K+1} - \bu_{ (m-1)K+k}\|_2 \leq P_T(m),
    \end{split}
\end{equation}
where $t = (m-1)K + k$.

Moreover,  the sum of all local path-length $ P_T(m) $ is upper bounded by the global path-length $ P_T $, i.e.,
\begin{equation}
    \label{eq:theorem_2_3}
    \sum_{m=1}^{T/K}  P_T(m) \leq  P_T.
\end{equation}
Therefore, substituting \eqref{eq:Lipschitz}, \eqref{eq:theorem_2_2} and \eqref{eq:theorem_2_3} into the third term $C$ of \eqref{eq:theorem_2}, we have
\begin{equation}
    \label{eq:theorem_2_4}
    \begin{split}
        C &= \sum_{m=1}^{T/K} \sum_{k=1}^K \left<\nabft(\bx_{m}), \bu_{s(m)} - \bu_{t} \right>  \\
        &\overset{\eqref{eq:Lipschitz}}{\leq}  G \sum_{m=1}^{T/K} \sum_{k=1}^K \| \bu_{s(m)} - \bu_t\|_2   \overset{\eqref{eq:theorem_2_2}}{\leq}  KG \sum_{m=1}^{T/K}  P_T(m) \overset{\eqref{eq:theorem_2_3}}{\leq}   KG P_T, 
    \end{split}
\end{equation}
where we denote $t = (m-1)K+k$ for brevity.

Combining \eqref{eq:theorem_2_1}, \eqref{eq:theorem_2_5}, \eqref{eq:theorem_2_4}, we have
\begin{equation}
    \label{eq:dynamic_BOGD}
    \begin{split}
        \text{{\rm D-Regret}}_T(\bu_1, \cdots, \bu_T)  &\leq GT \sqrt{3\epsilon} + \frac{7}{4 \eta} D^2 + \frac{\eta}{2}KTG^2 + \left(\frac{D}{\eta} + K G \right) P_T\\
        &=  (\sqrt{3}G + \frac{7}{4}D^2 + \frac{1}{2}G^2 + D P_T) T^{3/4} + G T^{1/2}  P_{T}  \\
        &= \mathcal{O}\left( T^{3/4} \left( 1 + P_T\right)\right),
    \end{split}
\end{equation}
where $\eta = T^{-3/4}$, $K = \eta^{-2/3} = T^{1/2}$, and  $\epsilon = \eta^{2/3} = T^{-1/2}$.

As shown in Lemma \ref{lemma:IP}, each invocation of $\O_{\text{IP}}$ requires solving at most $\O(T^{1/2})$ linear optimizations with above parameter choice. Therefore, the total number of solving linear optimizations is $N_{LOO} = \O(T)$, since $\text{BOGD}_{\text{IP}}$ only uses $T^{1/2}$ calls to $\O_{\text{IP}}$ with the block size $K = T^{1/2}$.

\subsection{Proof of Lemma \ref{lemma:IP}}

The proof can be found in \citet{COLT:2022:Garber}, but for the sake of completeness, we present it in detail here.

In $\text{BOGD}_\text{IP}$, at the end  of block $m-1$, we invoke $\O_{\text{IP}}$ as following:
\begin{equation}
    \bx_{m}, \Tilby_{m} =  \mathcal{O}_{\text{IP}}(\mathcal{K}, \epsilon, \bx_{m-1}, \by_{m}).
\end{equation}
According to Lemma \ref{lemma:lemma-IP}, we obtain that 
\begin{equation*}
     \|\bx_{m} - \Tilby_{m}\|_2 \leq \sqrt{3 \epsilon}, \quad \Tilby \in R\B.
\end{equation*}
And the number of calls to $\O_{\text{IP}}$ is
\begin{equation}
    n_{\text{IP}} = \max \left\{ \frac{ \|\bx_m - \by_{m+1}\|_2^2 (\|\bx_m - \by_{m+1}\|_2^2 - \epsilon)}{4 \epsilon^2}  + 1, 1 \right\}.
\end{equation}
According to Lemma \ref{lemma:FW-iterations}, each invocation on $\O_{\text{IP}}$ solves at most $\lceil \frac{27D^2}{4\epsilon} - 2  \rceil$ linear optimizations. Hence, the number of solving linear optimizations is 
\begin{equation}
    n_{\text{LO}} = \max \left\{ \frac{ \|\bx_m - \by_{m+1}\|_2^2 (\|\bx_m - \by_{m+1}\|_2^2 - \epsilon)}{4 \epsilon^2}  + 1, 1 \right\} \cdot \left \lceil \frac{27D^2}{4\epsilon} - 2 \right \rceil
\end{equation}

And by the update step $\by_{m+1} = \Tilby_{m} - \eta \sum_{r=(m-1)K+1}^{mK} \nabla  f_r(\bx_{m})$, we have
\begin{equation}
    \begin{split}
        \|\bx_m - \by_{m+1}\|_2 &=\left \|\bx_m - \Tilby_{m} + \eta \sum_{r=(m-1)K+1}^{mK}   \nabla  f_r(\bx_{m}) \right \|_2 \\
        &\overset{\eqref{eq:Lipschitz}}{\leq}  \|\bx_m - \Tilby_{m}\|_2 + \eta K G \overset{\eqref{eq:lemma-IP}}{\leq} \sqrt{3\epsilon} +  \eta K G.
    \end{split}
\end{equation}
Hence, according to $(a+b)^2 \leq 2a^2 + 2b^2$, we have
\begin{equation}
    \label{eq:NLOO-2}
    \|\bx_m - \by_{m+1}\|_2^2 \leq  (\sqrt{3 \epsilon} + \eta K G)^2 \leq 6\epsilon + 2\eta^2  K^2 G^2.
\end{equation}
Therefore, the number of solving linear optimizations is at most
\begin{equation}
    \begin{split}
       n_{\text{LO}}  &\leq  \left [\frac{ (6\epsilon + 2\eta^2  K^2 G^2) ( 6\epsilon + 2\eta^2  K^2 G^2 - \epsilon)}{4 \epsilon^2} + 1 \right] \cdot \left \lceil \frac{27D^2}{4\epsilon} - 2 \right \rceil  \\
        &\leq \left( 8.5 + 5.5 \frac{\eta^2 K^2 G^2}{\epsilon} + \frac{\eta^4 K^4 G^4}{\epsilon^2}\right)  \frac{27D^2}{4\epsilon} = \O(T^{1/2}), 
    \end{split}
\end{equation}
where  $\eta = T^{-3/4}$, $K = \eta^{-2/3} = T^{1/2}$ and $\epsilon = \eta^{2/3} = T^{-1/2}$.

\subsection{Proof of Lemma \ref{lemma:P_T}}
By the definition of $P_T'$, we have
\begin{equation}
    \begin{split}
        P_T' &= \sum_{m=2}^{T/K} \|  \bu_{s(m-1)} -  \bu_{s(m)} \|_2\\
        &= \|  \bu_{s(1)} -  \bu_{s(2)} \|_2 + \|  \bu_{s(2)} -  \bu_{s(3)} \|_2 + \cdots + \|  \bu_{s(T/K-1)} -  \bu_{s(T/K)} \|_2.
    \end{split}
\end{equation}
According to triangle inequality, we have
\begin{equation}
    \label{eq:lemma_P_T_1}
    \begin{split}
        \|  \bu_{s(1)} -  \bu_{s(2)} \|_2 &= \|  \bu_{1} -  \bu_{K+1} \|_2 \\ 
        &\leq \|  \bu_{1} -  \bu_{2} \|_2 + \cdots + \|  \bu_{K} -  \bu_{K+1} \|_2\\
        \|  \bu_{s(2)} -  \bu_{s(3)} \|_2 &= \|  \bu_{K+1} -  \bu_{2K+1} \|_2  \\
        &\leq \|  \bu_{K+1} -  \bu_{K+2} \|_2 + \cdots + \|  \bu_{2K} -  \bu_{2K+1} \|_2\\
        &\cdots \\
        \|  \bu_{s(T/K-1)} -  \bu_{s(T/K)} \|_2 &= \|  \bu_{T-2K+1} -  \bu_{T-K+1} \|_2  \\
        &\leq \|  \bu_{T-2K+1} -  \bu_{T-2K+2} \|_2 + \cdots + \|  \bu_{T-K} -  \bu_{T-K+1} \|_2.
    \end{split}
\end{equation}
Summing both side of \eqref{eq:lemma_P_T_1}, we have
\begin{equation}
    P_T' \leq \sum_{t=2}^{T-K+1} \|  \bu_{t-1} -  \bu_{t} \|_2 \leq \sum_{t=2}^{T} \|  \bu_{t-1} -  \bu_{t} \|_2 = P_T
\end{equation}

\subsection{Proof of Lemma \ref{lemma:BOGD_B}}
The analysis is inspired by \citet{ICML:2003:Zinkevich}, but we consider a more general case since Lemma \ref{lemma:BOGD_B} can nearly reduce to Theorem 2 in \citet{ICML:2003:Zinkevich} by setting the block size  $K=1$.

For brevity, we denote $t = (m-1)K+k$ and  $\bu_{s(m)}$ as the first action in the block $m$ where  $s(m) = (m-1)K+1$. In the block $m$, we have $\by_{m+1} = \Tilby_{m} - \eta \sum_{r=(m-1)K+1}^{mK} \nabla  f_{r}(\bx_{m}) = \Tilby_{m} - \eta\sum_{k=1}^K \nabft(\bx_m)$. Hence, we can prove that
\begin{equation}
    \begin{split}
        & \sum_{k=1}^K \left<\nabft(\bx_m), \Tilby_m - \bu_{s(m)} \right>  = \frac{1}{\eta}\left<  \Tilby_m - \by_{m+1},  \Tilby_m - \bu_{s(m)}\right> \\
        &= \frac{1}{2\eta} \left( \| \Tilby_m - \bu_{s(m)} \|_2^2 - \|  \by_{m+1} - \bu_{s(m)} \|_2^2 +  \| \Tilby_t -  \by_{t+1} \|_2^2 \right)\\
        &= \frac{1}{2\eta} \left( \| \Tilby_m - \bu_{s(m)} \|_2^2 - \|  \by_{m+1} - \bu_{s(m)} \|_2^2 \right) + \frac{\eta}{2} \| \sum_{k=1}^K \nabft(\bx_m)\|_2^2 \\
        &\overset{\eqref{eq:Lipschitz}}{\leq}\frac{1}{2\eta} \left( \| \Tilby_m - \bu_{s(m)} \|_2^2 - \|  \Tilby_{m+1} - \bu_{s(m)} \|_2^2 \right) + \frac{\eta}{2} K^2G^2\\
        &= \frac{1}{2\eta} \left( \| \Tilby_m\|_2^2 - \|  \Tilby_{m+1}\|_2^2 \right) + \frac{1}{\eta} \left< \Tilby_{m+1} -  \Tilby_m, \bu_{s(m)}\right> + \frac{\eta}{2}K^2 G^2.
    \end{split}
\end{equation}
Let $P_T'$ be  the path-length of the first action $\bu_{s(m)}$ in block $m$, i.e.,
\begin{equation}
    P_T' = \sum_{m=2}^{T/K} \|  \bu_{s(m-1)} -  \bu_{s(m)} \|_2.
\end{equation}
Summing up both side from $m=1$ to $T/K$, we have
\begin{equation}
    \begin{split}
        B &= \sum_{m=1}^{T/K} \sum_{k=1}^K \left<\nabft(\bx_m), \Tilby_m - \bu_{s(m)} \right> \\
        &\leq \frac{1}{2\eta}  \| \Tilby_1\|_2^2  + \frac{1}{\eta} \sum_{m=1}^{T/K}  \left< \Tilby_{m+1} -  \Tilby_m, \bu_{s(m)}\right> + \frac{\eta}{2}K T G^2 \\
        &\leq  \frac{1}{2\eta}  \| \Tilby_1\|_2^2 + \frac{1}{\eta} \left( \Tilby_{T/K+1}^\top \bu_{s(T/K)} - \Tilby_{1}^\top \bu_1\right) + \frac{1}{\eta}  \sum_{m=2}^{T/K} \left< \bu_{s(m-1)} -  \bu_{s(m)}, \Tilby_t\right>   + \frac{\eta}{2}K T G^2\\
        &\leq \frac{7}{4 \eta} D^2 + \frac{\eta}{2}KTG^2 + \frac{D}{\eta}P_T'
    \end{split}
\end{equation}
The last inequality is due to that $\forall t$, $\Tilby_t, \bu_t \in R\B$ ($D=2R$).

\subsection{Proof of Theorem \ref{theorem:Ader_BOGD}}
The key is to divide the dynamic regret into two terms, and upper bound them separately. 
First, we show that the dynamic regret can be decomposed as follows.
\begin{equation}
    \label{eq:theorem_3}
    \begin{split}
        \sum_{t=1}^T f_t(\bx_t) - \sum_{t=1}^T f_t(\bu_t) = \underbrace{\sum_{t=1}^T f_t(\bx_t) - \sum_{t=1}^T f_t(\bx_t^k)}_{:=A}+ \underbrace{\sum_{t=1}^T f_t(\bx_t^k) - \sum_{t=1}^T f_t(\bu_t)}_{:=B},
    \end{split}
\end{equation}
where $k = \lfloor \frac{3}{4}\log_2 (1+\frac{4P_T}{7D})\rfloor + 1$.

To upper bound the term A of \eqref{eq:theorem_3}, we introduce the following lemma.
\begin{lemma}
    (Lemma 1 in \citet{NeurIPS:2018:Zhang})
    Let $C = 1 + \frac{1}{N}$ and $w_1^i = \frac{C}{i(i+1)}$ for any expert $i$. Under Assumption \ref{assump:f-bound}, Algorithm \ref{alg3} satisfies
    \begin{equation}
        \sum_{t=1}^T f_t(\bx_t) - \sum_{t=1}^T f_t(\bx_t^i) \leq \frac{\sqrt{2T}}{4}[1 + 2\ln(i+1)].
    \end{equation}
\end{lemma}

And, the term $A$ of \eqref{eq:theorem_3} can be bounded as following
\begin{equation}
    A = \sum_{t=1}^T f_t(\bx_t) - \sum_{t=1}^T f_t(\bx_t^k) \leq  \frac{\sqrt{2T}}{4}[1 + 2\ln(k+1)],
\end{equation}
where  $k = \lfloor \frac{3}{4}\log_2 (1+\frac{4P_T}{7D})\rfloor + 1$.

Then, we proceed to upper bound the term $B$ of \eqref{eq:theorem_3}. According to Theorem \ref{theorem:dynamic_BOGD}, for any expert $i$, we have
\begin{equation}
    \label{eq:theorem_3_1}
    \begin{split}
        \sum_{t=1}^T f_t(\bx_t^i) - \sum_{t=1}^T f_t(\bu_t) &\leq GT \sqrt{3\epsilon_i} + \frac{7}{4 \eta_i} D^2 + \frac{\eta_i}{2}K_i TG^2 + \left(\frac{D}{\eta_i} + K_i G \right) P_T,
    \end{split}
\end{equation}
where $\epsilon_i = \eta_i^{2/3}$ and $K_i = \eta_i^{-2/3}$.

According to previous analysis, given a certain path-length $\Tilde{P}_T$, we can actually choose a certain step size $\Tilde{\eta} = \O(T^{-3/4}(1+\Tilde{P}_T)^{3/4})$, and achieve an improved bound. In the following, we introduce how to search this target step size $\Tilde{\eta}$ by maintaining multiple experts.

To facilitate computations, we assign the target step size as: 
\begin{equation}
    \label{eq:theorem_3_2}
    \Tileta = \left(\frac{7D^2 + 4DP_T}{2G^2 T} \right)^{3/4}.
\end{equation}

Next, we show that $\Tileta$ can actually be found by using  multiple experts with different $\eta \in \H$.

By the definition of $P_T$, we have
\begin{equation}
    0 \leq P_T = \sum_{t=2}^T \|\bu_{t-1} - \bu_{t}\|_2 \leq  TD.
\end{equation}
Hence, we have
\begin{equation}
    \left(\frac{7D^2}{2G^2 T}\right)^{3/4} \leq \Tileta \leq  \left(\frac{7D^2 + 4D^2T}{2G^2 T}\right)^{3/4}.
\end{equation}

It can be verified that 
\begin{equation}
    \min \H \leq \left(\frac{7D^2}{2G^2 T}\right)^{3/4} \quad \text{and}  \quad \left(\frac{7D^2 + 4D^2T}{2G^2 T}\right)^{3/4} \leq \max \H.
\end{equation}

Thus, there exists expert $k$ that satisfies
\begin{equation}
    \label{eq:theorem_3_3}
    \eta_k =  2^{k-1} \left(\frac{7D^2}{2G^2 T}\right)^{3/4} \leq \Tileta \leq 2\eta_k,
\end{equation}
where $k = \lfloor \frac{3}{4}\log_2 (1+\frac{4P_T}{7D})\rfloor + 1$.

For expert $k$, we have
\begin{equation}
    \begin{split}
        B &=  \sum_{t=1}^T f_t(\bx_t^k) - \sum_{t=1}^T f_t(\bu_t) \\
        &\leq   GT \sqrt{3\epsilon_k} + \frac{7}{4 \eta_k} D^2 + \frac{\eta_k}{2}K_k TG^2 + \left(\frac{D}{\eta_k} + K_k G \right) P_T \\
        &=  GT \sqrt{3  \eta_k^{2/3}} + \frac{7}{4 \eta_k} D^2 + \frac{\eta_k}{2} \eta_k^{-2/3} TG^2 + \left(\frac{D}{\eta_k} +  \eta_k^{-2/3} G \right) P_T \\
        &\overset{\eqref{eq:theorem_3_3}}{\leq}  GT \sqrt{3   \Tileta^{2/3}} + \frac{7}{2 \Tileta} D^2 + \frac{1}{2}  \Tileta^{1/3} TG^2 + \left(\frac{2D}{\Tileta} +  2^{2/3}\Tileta^{-2/3} G \right) P_T \\
        &\overset{\eqref{eq:theorem_3_2}}{\leq} (3G^{3/2}+2G^{1/2}) T^{3/4}(7D^2 + 4DP_T)^{1/4}   +4 G^2  T^{1/2} \frac{P_T}{(7D^2 + 4DP_T)^{1/2}}
    \end{split}
\end{equation}
Therefore, the dynamic regret of Algorithm \ref{alg3} is upper bounded by
\begin{equation}
    \begin{split}
        \sum_{t=1}^T f_t(\bx_t) - \sum_{t=1}^T f_t(\bu_t)
        &= \sum_{t=1}^T f_t(\bx_t) - \sum_{t=1}^T f_t(\bx_t^k) + \sum_{t=1}^T f_t(\bx_t^k) - \sum_{t=1}^T f_t(\bu_t) \\
        &\leq \frac{\sqrt{2T}}{4}[1 + 2\ln(k+1)] + (3G^{3/2}+2G^{1/2}) T^{3/4}(7D^2 + 4DP_T)^{1/4}  \\
        &+4 G^2  T^{1/2} \frac{P_T}{(7D^2 + 4DP_T)^{1/2}}   \\
        &= \mathcal{O}\left(T^{3/4}(1+P_T)^{1/4}\right).
    \end{split}
\end{equation}
where $k = \lfloor \frac{3}{4}\log_2 (1+\frac{4P_T}{7D})\rfloor + 1$.

\subsection{Proof of Theorem \ref{theorem:APOD-adaptive_regret}}
The analysis is divided into two parts. First, we upper bound the strongly adaptive regret of Algorithm \ref{alg-APOD} over any interval $J = [i, j] \in \mathcal{I}$. Then, we extend the regret bound over any interval $I = [s, s + \tau - 1] \subseteq [T]$.

For $J = [i, j] \in \mathcal{I}$, we have
\begin{equation}
    \label{eq:APOD-adaptive_expert}
    \begin{split}
        \sum_{u = i}^{j} f_u(\bx_{u, J}) - \min_{\bx \in \mathcal{K}} \sum_{u = i}^{j} f_u(\bx)
        &\overset{\eqref{eq:dynamic_BOGD}}{\leq} G|J| \sqrt{3\epsilon} + \frac{7}{4 \eta} D^2 + \frac{\eta}{2}K|J|G^2\\
        &\leq \left( \sqrt{3}G  + \frac{7}{4}D^2    + \frac{1}{2} G^2 \right)|J|^{3/4},
    \end{split}
\end{equation}
where the first inequality is due to \eqref{eq:dynamic_BOGD} with the static comparator (i.e., $P_T = 0$) and the second inequality is due to $\eta = |J|^{-3/4}, \epsilon = \eta^{2/3} = |J|^{-1/2}, K =  \eta^{-2/3}= |J|^{-1/2}$.

First, we present the regret bound of meta algorithm (POLA) with respect to experts ($\text{BOGD}_{\text{IP}}$).
\begin{lemma}
    \label{lemma:APOD-interval}
    (Lemma 8 in \citet{AISTATS:2020:Zhang}) Under Assumption \ref{assump:f-bound}, for any interval $J = [i, j] \in \I$, Algorithm \ref{alg-APOD} guarantees
    \begin{equation}
        \label{eq:APOD-adaptive_meta}
        \sum_{u = i}^{j} f_u(\bx_u) - \sum_{u = i}^{j} f_u(\bx_{u, J}) \leq \sqrt{3 c(j) |J| },
    \end{equation}
    where $c(j)\leq 1 + \ln j + \ln(1 + \log_2 j) + \ln\frac{5+3\ln(1+j)}{2}$.
\end{lemma}

Then, combining \eqref{eq:APOD-adaptive_meta} and \eqref{eq:APOD-adaptive_expert}, we can obtain that
\begin{equation}
    \label{eq:APOD-adaptive_regret}
    \sum_{u = i}^{j} f_u(\bx_u)  - \min_{\bx \in \mathcal{K}} \sum_{u = i}^{j} f_u(\bx)
    \leq \sqrt{3  c(j)}  |J|^{1/2} + \left( \sqrt{3}G  + \frac{7}{4}D^2    + \frac{1}{2} G^2 \right)|J|^{3/4}.
\end{equation}

Now, we proceed to extend the regret bound over $J$ to any interval $I = [s, s + \tau - 1] \subseteq [T]$. The key of the extension is that  $I = [s, s + \tau - 1]$ can be partitioned into two sequences of intervals in GC, as shown below.
\begin{lemma}
    \label{lemma:interval_extension}
    (Lemma 1.2 of \citet{ICML:2015:Daniely}) Any interval $I = [s, s + \tau - 1] \subseteq [T]$ can be into two sequences of disjoint and consecutive intervals, $I_{-p}, \cdots, I_0 \in \I$ and $I_{1}, \cdots, I_q \in \I$, which satisfy
    \begin{equation}
        \label{eq:interval_extension}
        \begin{split}
            \forall i \geq 1, |I_{-i}|/|I_{-i+1}| \leq 1/2 \quad \text{and} \quad \forall i \geq 2, |I_{i}|/|I_{i-1}| \leq 1/2
        \end{split}
    \end{equation}

\end{lemma}

According to Lemma \ref{lemma:interval_extension}, for any fixed $\bx \in \mathcal{K}$, we can obtain that
\begin{equation}
    \begin{split}
        &\sum_{t = s}^{s + \tau - 1}  f_t(\bx_t) - \sum_{t = s}^{s + \tau - 1}  f_t(\bx) \\
        &= \sum_{i = -p}^q \left(\sum_{t \in I_i}  f_t(\bx_t) - \sum_{t \in I_i} f_t(\bx) \right) \\
        &\overset{\eqref{eq:APOD-adaptive_regret}}{\leq} \sum_{i = -p}^q \left( \sqrt{3  c(s + \tau - 1)}  |I_i|^{1/2}+ \left( \sqrt{3}G  + \frac{7}{4}D^2    + \frac{1}{2} G^2 \right)|I_i|^{3/4}  \right)\\
        &\overset{\eqref{eq:interval_extension}}{\leq}  2\sqrt{3  c(s + \tau - 1) } \sum_{i=0}^{\infty}(2^{-i}\tau)^{1/2} + 2\left( \sqrt{3}G  + \frac{7}{4}D^2 + \frac{1}{2} G^2 \right) \sum_{i=0}^{\infty}(2^{-i}\tau)^{3/4}\\
        &\leq 8\sqrt{3  c(s + \tau - 1) } \tau^{1/2} + 6 \left( \sqrt{3}G  + \frac{7}{4}D^2 + \frac{1}{2} G^2 \right) \tau^{3/4}.
    \end{split}
\end{equation}

Therefore, the strongly adaptive regret bound of Algorithm \ref{alg-APOD} is
\begin{equation}
    \begin{split}
        \text{{\rm SA-R}}(T, \tau)
        &= \max_{[s, s + \tau - 1] \subseteq [T]} \left(  \sum_{t = s}^{s + \tau - 1}  f_t(\bx_t) - \min_{\bx \in \mathcal{K}} \sum_{t = s}^{s + \tau - 1}  f_t(\bx) \right) \\
        &\leq 8\sqrt{3  c(T) } \tau^{1/2} + 6 \left( \sqrt{3}G  + \frac{7}{4}D^2 + \frac{1}{2} G^2 \right) \tau^{3/4} = \Tilde{\O}\left( \tau^{3/4}\right),
    \end{split}
\end{equation}
where  $c(T)\leq 1 + \ln T + \ln(1 + \log_2T) + \ln\frac{5+3\ln(1+T)}{2}$.

\subsection{Proof of Theorem \ref{theorem:APOD-dynamic_regret}}
The analysis is similar to \citet{AISTATS:2020:Zhang}, and the key is to prove the dynamic regret bound on the experts $\{E_{I_k^1}, E_{I_k^2}, \cdots\}$ over several interval sets $\{I_k^1, I_k^2, \cdots\} \subseteq \I$.

Specifically, due to
\begin{equation}
    P_T = \sum_{t=2}^T \|\bu_{t-1} - \bu_t\|_2 \in [0, DT],
\end{equation}
we divide the path-length $P_T$ into two cases:  $P_T \in [0, D]$ and $P_T \in (D, DT]$, and establish the dynamic regret in the two cases separately.

\subsubsection{Case 1: \texorpdfstring{$P_T \in [0, D]$} .}
Note that
\begin{equation}
    \I = \I_0 \cup \I_1 \cup \cdots  \cup \I_{\alpha}.
\end{equation}
For brevity, we denote $\alpha = \lfloor\log_2 T\rfloor$ and $I_j^1 = [2^j, 2^{j+1}-1]$ as the  first interval of  $\I_j \subseteq \{\I_0, \cdots, \I_\alpha\}$.
We upper bound the dynamic regret over $I_j^1 (j = 0, \cdots, \alpha)$, as shown below
\begin{equation}
    \begin{split}
        \sum_{t=1}^T f_t(\bx_t) - \sum_{t=1}^T f_t(\bu_t)
        &= \sum_{j=0}^{\alpha-1} \left\{  \sum_{t=2^{j}}^{2^{j+1}-1} \Big( f_t(\bx_t) -  f_t(\bu_{t})\Big)  \right\}
        +  \sum_{t=2^\alpha}^{T} \Big( f_t(\bx_t) -  f_t(\bu_{t})\Big)\\
        &= \underbrace{\sum_{j=0}^{\alpha-1} \left\{  \sum_{t=2^{j}}^{2^{j+1}-1}  \Big( f_t(\bx_t) -  f_t(\bx_{t,I_j^1})\Big)  \right\} + \sum_{t=2^\alpha}^{T} \Big( f_t(\bx_t) -  f_t(\bx_{t,I_\alpha^1})\Big)}_{:=A} \\
        &+ \underbrace{\sum_{j=0}^{\alpha-1} \left\{  \sum_{t=2^{j}}^{2^{j+1}-1}   \Big( f_t(\bx_{t,I_j^1}) -  f_t(\bu_t)\Big)  \right\}   + \sum_{t=2^\alpha}^{T} \Big( f_t(\bx_{t,I_\alpha^1}) -  f_t(\bu_t)\Big)}_{:=B}\\
    \end{split}
\end{equation}

We proceed to bound the term $A$. By using Lemma \ref{lemma:APOD-interval}, we have
\begin{equation}
    \label{APOD-dynamic_regret-1-1}
    \begin{split}
        A
        &\overset{\eqref{eq:APOD-adaptive_meta}}{\leq} \sum_{j=0}^{\alpha-1}  \sqrt{3c(2^{j+1}-1) 2^{j-1}} + \sqrt{3c(T) (T - 2^\alpha)} \\
        &\leq \sqrt{27c(T)  2^\alpha} + \sqrt{3c(T) (T - 2^\alpha)}\\
        &\leq \sqrt{30c(T) T},
    \end{split}
\end{equation}
where the second inequality is due to $\forall j\leq \alpha - 1, 2^{j+1} - 1 \leq T$ , and the third inequality is due to Cauchy-Schwarz Inequality.

For simplicity, we denote $P_{i:j} = \sum_{t=i+1}^j \|\bu_{t-1} - \bu_t \|_2$ and the path-length $P_T$ can be decomposed as following:
\begin{equation}
    P_T = P_{2^0:(2^1-1)} + \cdots + P_{2^j:(2^{j+1}-1)} + \cdots + P_{ 2^\alpha :T}
\end{equation}

Then, we proceed to bound the term $B$.
\begin{equation}
    \label{APOD-dynamic_regret-1-2}
    \begin{split}
        B
        &\overset{\eqref{eq:dynamic_BOGD}}{\leq} \sum_{j=0}^{\alpha-1} \left\{ (\sqrt{3}G + \frac{7}{4}D^2 + \frac{1}{2}G^2) \left(2^{j} \right)^{3/4} + D  \left(2^{j} \right)^{3/4}  P_{2^j:(2^{j+1}-1)} + G\left(2^{j} \right)^{1/2}   P_{2^j:(2^{j+1}-1)}  \right\}\\
        &+ (\sqrt{3}G + \frac{7}{4}D^2 + \frac{1}{2}G^2) \left(T - 2^\alpha \right)^{3/4} + D \left(T - 2^\alpha \right)^{3/4}  P_{ 2^\alpha :T} + G\left(T - 2^\alpha\right)^{1/2}  P_{ 2^\alpha :T}\\
        &\leq 3(\sqrt{3}G + \frac{7}{4}D^2 + \frac{1}{2}G^2) \left(2^{\alpha} \right)^{3/4} + D  \left(2^{\alpha} \right)^{3/4} P_{1:(2^{\alpha}-1)} + G\left(2^{\alpha} \right)^{1/2}  P_{1:(2^{\alpha}-1)}   \\
        &+ (\sqrt{3}G + \frac{7}{4}D^2 + \frac{1}{2}G^2) \left(T - 2^\alpha \right)^{3/4} + D \left(T - 2^\alpha \right)^{3/4}  P_{ 2^\alpha :T} + G\left(T - 2^\alpha\right)^{1/2}  P_{ 2^\alpha :T}\\
        &\leq  4(2\sqrt{3}G + \frac{7}{2}D^2 + G^2)  T^{3/4} + 2D^{7/4}  T^{3/4} P_{T}^{1/4} + 2G D^{1/2}T^{1/2}  P_{T}^{1/2},
    \end{split}
\end{equation}
where  the last inequality is due to $\frac{T}{2} \leq 2^\alpha \leq T$ and $P_T \leq D$.

We complete the proof by combining \eqref{APOD-dynamic_regret-1-1}  and \eqref{APOD-dynamic_regret-1-2}.

\subsubsection{Case 2: \texorpdfstring{$P_T \in (D, DT]$} .}
For this case, we divide $(D, DT]$ as following
\begin{equation}
    \underbrace{(D2^{0}, D2^1]}_{\delta_1}, \quad\underbrace{(D2^{1}, D2^2]}_{\delta_2}, \quad\cdots, \quad\underbrace{(D2^{i-1}, D2^i]}_{\delta_i}, \quad\cdots, \quad\underbrace{(D2^{s-1}, D2^s]}_{\delta_s},
\end{equation}
where $s = \lfloor \log T\rfloor$.
And for the interval $\delta_i = (D2^{i-1}, D2^i]$, we analyze the dynamic regret bound over  $\I_0, \cdots, \I_{s-i}$. Specifically, we consider the first interval in $\I_0, \cdots, \I_{s-i-1}$, i.e.,~$I_0^1, \cdots, I_{s-i-1}^1$ and all intervals in $\I_{s-i}$, i.e.,~$I_{s-i}^1, \cdots, I_{s-i}^u,\cdots, I_{s-i}^m ~(m = \lceil T/2^{s-i}\rceil - 1)$. For brevity, the beginning of $I_{s-i}^u$ is denoted as $s^u = u \cdot 2^{s-i}$ and the end of  $I_{s-i}^u$ is denoted as $e^u = (u+1) \cdot 2^{s-i}-1$.

Besides, by the fact that
\begin{equation}
    D2^{i-1}\leq P_T \leq D2^i, \quad and \quad s =  \lfloor \log T\rfloor,
\end{equation}
we have
\begin{equation}
    \label{eq:interval_ineq}
    2^{s} \leq T, \quad 2^{s-i} \leq T, \quad \frac{P_T}{D} \leq \frac{T}{2^{s-i}} \leq \frac{2P_T}{D}.
\end{equation}

Similar to the case 1, the dynamic regret can be decomposed as below.
\begin{equation}
    \label{eq:APOD_case_1_dynamic}
    \begin{split}
        &\sum_{t=1}^T f_t(\bx_t) - \sum_{t=1}^T f_t(\bu_t) \\
        &= \sum_{j=0}^{s-i-1} \left\{ \sum_{t=2^{j}}^{2^{j+1}-1} \left(f_t(\bx_t) -  f_t(\bu_{t})\right) \right\} + \sum_{u=1}^{m-1} \left( \sum_{t=s^u}^{e^u} \big(f_t(\bx_t) -  f_t(\bu_{t})\big)\right) + \sum_{t=s^m}^{T} \Big( f_t(\bx_t) -  f_t(\bu_{t})\Big) \\
        &= \sum_{j=0}^{s-i-1} \left\{ \sum_{t=2^{j}}^{2^{j+1}-1} \left(f_t(\bx_t) -  f_t(\bx_{t, I_{j}^1})\right) \right\} + \sum_{u=1}^{m-1} \left( \sum_{t=s^u}^{e^u} \big(f_t(\bx_t) -  f_t(\bx_{t, I_{s-i}^m})\big)\right)\\
        &+ \sum_{t=s^m}^{T} \Big( f_t(\bx_t) -  f_t(\bx_{t, I_{s-i}^m})\Big)
        +\sum_{j=0}^{s-i-1} \left\{ \sum_{t=2^{j}}^{2^{j+1}-1} \left(f_t(\bx_{t, I_{j}^1}) -  f_t(\bu_{t})\right) \right\}\\
        &+ \sum_{u=1}^{m-1} \left( \sum_{t=s^u}^{e^u} \big(f_t(\bx_{t, I_{s-i}^m}) -  f_t(\bu_{t})\big)\right) + \sum_{t=s^m}^{T} \Big( f_t(\bx_{t, I_{s-i}^m}) -  f_t(\bu_{t})\Big).
    \end{split}
\end{equation}
To simplify the notation, we denote \eqref{eq:APOD_case_1_dynamic} as the sum of term $A$ and term $B$, where  term $A$ is defined as 
\begin{equation}
    \begin{split}
        A &:=  \sum_{j=0}^{s-i-1} \left\{ \sum_{t=2^{j}}^{2^{j+1}-1} \left(f_t(\bx_{t, I_{j}^1}) -  f_t(\bu_{t})\right) \right\} + \sum_{u=1}^{m-1} \left( \sum_{t=s^u}^{e^u} \big(f_t(\bx_{t, I_{s-i}^m}) -  f_t(\bu_{t})\big)\right) \\
        &+ \sum_{t=s^m}^{T} \Big( f_t(\bx_{t, I_{s-i}^m}) -  f_t(\bu_{t})\Big), 
    \end{split}
\end{equation}
and term $B$ is defined as 
\begin{equation}
    \begin{split}
        B &:=  \sum_{j=0}^{s-i-1} \left\{ \sum_{t=2^{j}}^{2^{j+1}-1} \left(f_t(\bx_t) -  f_t(\bx_{t, I_{j}^1})\right) \right\} + \sum_{u=1}^{m-1} \left( \sum_{t=s^u}^{e^u} \big(f_t(\bx_t) -  f_t(\bx_{t, I_{s-i}^m})\big)\right)\\
        &+ \sum_{t=s^m}^{T} \Big( f_t(\bx_t) -  f_t(\bx_{t, I_{s-i}^m})\Big), 
    \end{split}
\end{equation}

In the following, we first upper bound the term $A$. By using Lemma \ref{lemma:APOD-interval}, we have
\begin{equation}
    \label{eq:APOD_case_2_dynamic_A}
    \begin{split}
        A
        &\overset{\eqref{eq:APOD-adaptive_meta}}{\leq} \sum_{j=0}^{s-i-1} \sqrt{3 c(2^{j+1}-1)2^j}  + \sum_{u=1}^{m-1} \sqrt{3 c(e^u) (e^u - s^u + 1)} + \sqrt{3 c(T) (T-s^m+1)} \\
        &\leq \sqrt{12 c(T) s^1}   + \sum_{u=1}^{m-1} \sqrt{12 c(T) (e^u - s^u + 1)} + \sqrt{12 c(T) (T-s^m+1)} \\
        &\leq \sqrt{12(m+1) c(T) T}  \leq 5\sqrt{c(T) T \left(1 + \frac{P_T}{D} \right)},
    \end{split}
\end{equation}
where the last inequality is due to that
\begin{equation}
    m \leq 1 + \frac{T}{2^{s-i}} \overset{\eqref{eq:interval_ineq}}{\leq}  1 + \frac{2P_T}{D}.
\end{equation}

Then, we upper bound the term B.
\begin{equation}
    \label{eq:APOD_case_2_dynamic_B}
    \begin{split}
        B
        &\overset{\eqref{eq:dynamic_BOGD}}{\leq} \sum_{j=0}^{s-i-1} \left\{ (\sqrt{3}G + \frac{7}{4}D^2 + \frac{1}{2}G^2) \left(2^j\right)^{3/4} + D  \left(2^j\right)^{3/4} P_{2^j:(2^{j+1}-1)} + G \left(2^j\right)^{1/2}  P_{2^j:(2^{j+1}-1)} \right\}\\
        &+ \sum_{u=1}^{m-1} \left(  (\sqrt{3}G + \frac{7}{4}D^2 + \frac{1}{2}G^2) \left(e^u-s^u+1\right) ^{3/4} + D  \left(e^u-s^u+1\right) ^{3/4} P_{s^u:e^u} \right) \\
        &+ \sum_{u=1}^{m-1} \left( G \left(e^u-s^u+1\right) ^{1/2}  P_{s^u:e^u} \right) +  (\sqrt{3}G + \frac{7}{4}D^2 + \frac{1}{2}G^2) \left(T-s^m+1\right) ^{3/4}  \\
        &+ D  \left(T-s^m+1\right) ^{3/4} P_{s^m:T} + G \left(T-s^m+1\right) ^{1/2}  P_{s^m:T}\\
        &\leq 2(\sqrt{3}G + \frac{7}{4}D^2 + \frac{1}{2}G^2) \left(2^{s-i}\right)^{3/4} + D \left(2^{s-i}\right)^{3/4} P_{1:e^0} + G\left(2^{s-i}\right)^{1/2} P_{1:e^0}\\
        &+ \sum_{u=1}^{m-1} \left( 2 (\sqrt{3}G + \frac{7}{4}D^2 + \frac{1}{2}G^2) \left(2^{s-i}\right) ^{3/4} + D  \left(2^{s-i}\right) ^{3/4} P_{s^u:e^u} + G \left(2^{s-i}\right) ^{1/2}  P_{s^u:e^u}  \right)\\
        &+  2(\sqrt{3}G + \frac{7}{4}D^2 + \frac{1}{2}G^2) \left(2^{s-i}\right) ^{3/4} + D  \left(2^{s-i}\right) ^{3/4} P_{s^m:T} + G \left(2^{s-i}\right) ^{1/2}  P_{s^m:T}\\
        &\leq 2\left( \sqrt{3}G + \frac{7}{4}D^2 +  \frac{1}{2} G^2 \right) (m+1) \left(2^{s-i}\right)^{3/4} + D \left(2^{s-i}\right)^{3/4} P_T + G \left(2^{s-i}\right)^{1/2} P_T.
    \end{split}
\end{equation}
Note the fact that
\begin{equation}
    \label{eq:interval_ineq_2}
    m \leq 1 + \frac{T}{2^{s-i}}, \quad   2^{s-i} \leq T   \quad   \text{and} \quad  \frac{D}{2P_T}T \overset{\eqref{eq:interval_ineq}}{\leq} 2^{s-i} \overset{\eqref{eq:interval_ineq}}{\leq} \frac{D}{P_T}T.
\end{equation}
Substituting \eqref{eq:interval_ineq_2} into \eqref{eq:APOD_case_2_dynamic_B}, we have
\begin{equation}
    \label{eq:APOD_case_2_dynamic_B_2}
    \begin{split}
        B
        &\leq 4\left( \sqrt{3}G + \frac{7}{4}D^2 +  \frac{1}{2} G^2 \right) T^{3/4}
        + 4\left( \sqrt{3}G  D^{-1/4} + \frac{7}{4} D^{7/4} +  \frac{1}{2} G^2 D^{-1/4} \right)T^{3/4}P_T^{1/4}\\
        &+ D^{7/4} T^{3/4}P_T^{1/4} +  G D^{1/2} T^{1/2} P_T^{1/2}.
    \end{split}
\end{equation}
We complete the proof by combining \eqref{eq:APOD_case_2_dynamic_A} and \eqref{eq:APOD_case_2_dynamic_B_2}.


\end{document}